\documentclass[12pt]{mytmp}

\usepackage{fullpage}

\usepackage[utf8]{inputenc} 
\usepackage[T1]{fontenc}    
\usepackage{hyperref}       
\usepackage{url}            
\usepackage{booktabs}       
\usepackage{amsfonts}       
\usepackage{nicefrac}       
\usepackage{microtype}      
\usepackage{color}
\usepackage{graphicx}
\usepackage{subfigure} 
\usepackage{multirow}
\usepackage{amsmath}
\usepackage{amssymb, bm}
\usepackage{tablefootnote}

\def\AA{\mathcal{A}}

\def\OO{\mathcal{O}}
\def\PP{\mathcal{P}}
\def\SS{\mathcal{S}}

\begin{document}

\begin{frontmatter}

\title{TStarBot-X: An Open-Sourced and Comprehensive Study for Efficient League Training in StarCraft II Full Game}

\author[x]{Lei Han\corref{cor1}}
\ead{leihan.cs@gmail.com}
\author[x]{Jiechao Xiong\corref{cor1}}
\ead{jchxiong@gmail.com}
\author[x]{Peng Sun\corref{cor1}}
\ead{pengsun000@gmail.com}
\author[x]{Xinghai Sun}
\author[x]{Meng Fang}
\author[aipt]{Qingwei Guo}
\author[aipt]{Qiaobo Chen}
\author[aipt]{Tengfei Shi}
\author[aipt]{Hongsheng Yu}
\author[aipt]{Xipeng Wu}
\author[x]{Zhengyou Zhang}

\cortext[cor1]{Equal contribution. Correspondence to the first three authors.}

\address[x]{Tencent Robtics X, Shenzhen, China}
\address[aipt]{Department of AI Platform, Tencent, Shenzhen, China}

\begin{abstract}
StarCraft, one of the most difficult esport games with long-standing history of professional tournaments, 
has attracted generations of players and fans, and also, intense attentions in artificial 
intelligence research. Recently, Google's DeepMind announced AlphaStar~\cite{astar}, a grandmaster level AI 
in StarCraft II that can play with humans using comparable action space and operations. 
In this paper, we introduce a new AI agent, named \mbox{TStarBot-X}, that is trained under orders of less 
computations and can play competitively with expert human players 
\footnote{We focus on playing Zerg, one of the three StarCraft races. We need to clarify that we 
are not able to precisely place the agent at an official level such as Grandmaster, Master or Diamond, etc., 
since we can not access Battle.net League system through public APIs without Blizzard Entertainment's permission. 
Instead, we let the agent play against officially ranked
Grandmaster and Master level human players, and receive evaluations from humans. Please refer to \ref{sec:res} for the results.}. 
TStarBot-X takes advantage of important techniques introduced in AlphaStar, 
and also benefits from substantial innovations including new league training methods, novel multi-agent roles, 
rule-guided policy search, stabilized policy improvement, lightweight neural network architecture, 
and importance sampling in imitation learning, etc. 
We show that with orders of less computation scale, a faithful reimplementation of AlphaStar's methods can not succeed 
and the proposed techniques are necessary to ensure TStarBot-X's competitive performance. 
We reveal all technical details that are complementary to those mentioned in AlphaStar, showing the 
most sensitive parts in league training, reinforcement learning and imitation learning that 
affect the performance of the agents. Most importantly, this is an open-sourced study 
that all codes and resources (including the trained model parameters) are publicly accessible via 
\url{https://github.com/tencent-ailab/tleague_projpage}.
We expect this study could be beneficial for both academic and industrial future research in 
solving complex problems like StarCraft, and also, might provide a sparring partner for all StarCraft II players 
and other AI agents.

\end{abstract}

\begin{keyword}
Game \sep
StarCraft \sep 
Reinforcement Learning \sep 
Multi-Agent \sep 
League Training


\end{keyword}
\end{frontmatter}

\section{Introduction}
\label{sec:intro}

{G}{ame}, 
a naturally closed environment with well designed logic, has been considered as the most straightforward 
and ideal scenarios for testing artificial intelligence (AI). Recent advances in large-scale deep reinforcement 
learning (RL) have demonstrated that current learning systems are able to solve
complex games even with super-human performance. There have been a series of breakthroughs, including solving games of 
Atari~\cite{mnih2015human}, GO~\cite{silver2016mastering, silver2017mastering}, Quake III Arena Capture the Flag~\cite{jaderberg2019human} 
and DOTA2~\cite{openai-five}, etc. Real-time strategy (RTS) games, a sub-genre of video games that requires 
consistent timely decisions on a large amount of choices like resource-gathering, construction and units control, 
were considered as the most challenging benchmark for AI research. Recently, Google's DeepMind announced AlphaStar~\cite{astar}, a 
grandmaster level performance AI that can play in a way comparable to humans in StarCraft II (SC2), an iconic
representative of RTS game developed by Blizzard Entertainment. 
The AlphaStar's system turns to be an enormous infrastructure with its scopes covering advanced deep neural 
networks, imitation learning (IL), RL, multi-agent population-based training, huge-scale distributed actor-learner architecture, 
environmental engineering, etc. Even a faithful reimplementation of this system is nontrivial without
the same scale of computations and codes.

With full respect on the work of AlphaStar, we are curious about every detail how well the AI 
understands the complex logics in SC2, how each part in the learning system contributes to generating 
strong AIs, and how general these techniques can be applied especially when the computation resources are limited. 
In this paper, we try to answer these questions by introducing and analyzing a new AI agent, TStarBot-X, a 
successor of our previous work TStarBot1 and TStarBot2~\cite{sun2018tstarbots}. TStarBot-X differs much from its early 
versions (TStarBot1 and TStarBot2) which are based either on macro hierarchical actions or hand-craft rules~\cite{sun2018tstarbots}. 
It uses similar action space as AlphaStar and is comparable to human operations. 
The average APM/EPM of TStarBot-X is 232/196 
and the peak value is 609/519
(calculated from the matches in human evaluation), which is less than many expert human players and AlphaStar's APM/EPM.
\mbox{TStarBot-X} uses only 20 million 
parameters in its neural network, including both the policy net (with 17.25 million parameters) 
and the value net (with shared parts from policy and 2.75 million additional parameters), 
while the number of parameters used by AlphaStar is 139 million in total, with 55 million for policy net 
and 84 million for value net, respectively.
As we can clearly see that our model is much smaller than that of AlphaStar, especially for the value net. 
Nevertheless, the network seems to work well in performance and gains considerable efficiency.
As has been demonstrated in AlphaStar, imitation/supervised learning is very important to provide the agent
a good initial policy. TStarBot-X also starts learning by imitating human replays. We used a dataset 
containing 309,571 Zerg vs. Zerg replays across SC2 versions 4.8.2, 4.8.4, 4.8.6 and 4.9.0-4.9.3, where 
at least one player's MMR $\geq$ 3500.
The supervised agent can defeat the built-in Elite-bot (level 7) with a win-rate of 90\% with 
either the zero $z$-stat or sampling from a subset of 174 $z$-stats collected from human replays.
The $z$-stat is a high-level feature including strategic statistics introduced in~\cite{astar}.
It is worth mentioning that in IL, importance sampling over trajectories is very important to the agent performance.
Without importance sampling, i.e., using the original data from human replays, the win-rate against the Elite-bot only reaches 68\%.

After obtaining the supervised agent, we indeed immediately tried to follow AlphaStar's configuration to 
implement multi-agent reinforcement learning by populating the league~\cite{astar,jaderberg2019human} under 
our computational scale whose total consuming and generating speeds are only 1/30 and 1/73 of those in AlphaStar.
That is, we construct the league with one main agent (MA), two main exploiters (ME) and two league exploiters (LE).
\footnote{In AlphaStar's league, there are three main agents each of which is for one StarCraft race. Each main agent is 
equipped with one main exploiter and two league exploiters. To align with such a setting and make full use of our 
computational resources, we use one main agent, two main exploiters and two league exploiters for AlphaStar Surrogate. 
Detailed configurations can be found in Section \ref{sec:res}.}
We refer to this configuration as `AlphaStar Surrogate' that will be referenced as a baseline method.
Unfortunately, although the league training can consistently improve the agent's performance, 
the strategic diversity and strength in the entire league is significantly limited.
In several repetitions, the main agent converges to either Zergingling Rush or Roach Push (two early Zerg strategies)
in a few days, and the unit/upgrade types produced in the entire league are also very limited. 
We conjecture that insufficient explorations prevent the agents from improving as expected. 

In order to enhance the agent diversity and strength in the league, we propose three new techniques: 
Diversified League Training (DLT), Rule-Guided Policy Search (RGPS), and stabilized policy improvement with 
Divergence-Augmented Policy Optimization (DAPO)~\cite{wang2019divergence}. 
We need to first clarify that when we refer to diversity of the league, we are talking about the 
the diversity of the exploiters and diversity of the MA separately. The former indicates the diverse strategies generated
in the entire league, while the latter indicates the diversity of the counter-strategies and robustness 
when the MA plays against any exploiter using arbitrary strategies. 
The strength of the agents indicates that for a specific strategy how strong and well the agent can perform.

DLT thus aims to efficiently generate diverse and strong exploiters under our case. To do so, we first fine-tune 
the supervised agent on $K$ separate sub-datasets, in each of which some specific units or strategies take place 
in human replays, to obtain $K$ specific and diverse supervised agents. The $K$ sub-datasets can be simply divided
by unit types or summarized strategies by human experts. 
We then introduce several new agent roles in the league, including specific exploiter (SE), evolutionary exploiter 
(EE) and adaptive evolutionary exploiter (AEE). SE works in the same way as ME except that it starts with one of the above
$K$ fine-tuned supervised models. EE and AEE focus on continuing training historical exploiter agents that are still 
exploitable. EE tends to evolve as strong as possible probably with some distinct strategies, while AEE 
tends to balance the diversity and strength of the agents.
The league training is then launched by consisting of a mixture of MA, ME, LE, SE, EE and AEE.
It turns out that DLT can efficiently generate various units and strategies among the agents in the league (Figure~\ref{fig:diversity_oppo}).

DLT pushes much burden on MA's improvement. That is, MA needs to find all counter-strategies against diverse exploiters
and to be as robust as possible.
To speedup this progress, RGPS is introduced to help MA to skip tedious explorations on inherent game logics and let MA focus
on explorations at a strategic level. We use the following example to illustrate the aforementioned inherent game logics. 
For SC2 human players, they understand `Hydralisk' is strong against `Mutalisk' and its training relies on `Hydraliskden' 
which further relies on `Lair', by just reading the official player's instructions and quickly verifying it in a few practical matches.
\footnote{Please refer to \ref{sec:sc2} for explanations of these SC2 terms.}
However, it is extremely difficult for AI agents to understand these logics from a semantic level, and agents need to explore
with extremely large amount of data to fit such game logics into a parameterized policy.
Specifically, RGPS uses a distillation term to induce the policy to follow some inherent game logics 
conditioning on a few critical states, on which the teacher policy outputs an one-hot or multi-hot distribution that is a distributional
encoding of `if-else' logics. We show that with RGPS, MA can easily learn these game logics and focus on explorations on 
long-term strategies (Figure~\ref{fig:diversity_ma}).

With sufficient league diversity, we also consider to speedup the policy improvement using DAPO. DAPO is originally derived for 
stabilized policy optimization from the scope of mirror descent algorithms~\cite{wang2019divergence}. 
In league training, we apply DAPO at the agent level that the agent at learning period $t+1$ has to stay close to its 
historical model at learning period $t$ via a KL term. In our case, each agent stores a copy of itself into the league every 
$\sim$12 hours which is referred to as a learning period. In the formal experiment that lasts for
57 days, we use DLT+RGPS in the first 42 days and activate DAPO in the last 15 days where we observe significant policy improvement speedup
after activating DAPO (Figure~\ref{fig:elo}).

After 57 days of training, the final TStarBot-X agents are evaluated by four invited expert human players with tens of games 
(Table~\ref{tab:res-hum}). TStarBot-X wins two Master players with 11:0 and 13:0, and wins two Grandmaster players
with 7:0 and 4:1, respectively. Other comprehensive league evaluations are provided in the experimental section.
We end up the introduction with an overall comparison between AlphaStar's learning systems and ours in Table~\ref{tab:com}.

\begin{table*}[t]
  \centering
  \caption{A comparison between AlphaStar and TStarBot-X's learning systems. All numbers for AlphaStar are calculated from the reported numbers in \cite{astar} and converted to the same criterion for comparison. `fps' indicates `frame/step per second'. 
}
  \label{tab:com}
  \resizebox{\textwidth}{38mm}
  {
  \begin{tabular}{c|c|c}
    \hline
    \hline
     & \textbf{AlphaStar} & \textbf{TStarBot-X} \\
    \hline 
    \hline
	\textbf{Race} & All three SC2 races & Zerg \\
	\hline
	\multirow{2}{*}{\textbf{Map}} & Cyber Forest, Kairos Junction, & \multirow{2}{*}{Kairos Junction} \\
	 & King’s Cove, and New Repugnancy & \\
	\hline
    \textbf{Action Space} & Similar to human operations & Similar to human operations \\
    \hline
    \textbf{Use Camera View} & False for demonstration; True for final version & False \\
    \hline
    \textbf{Average APM/EPM} & 259/211 for Zerg & 232/196  \\
    \hline
    \textbf{Max APM/EPM} & 1192/823 for Zerg & 609/519  \\
    \hline
    \textbf{Model Size} & 139 million parameters & 20 million parameters \\
	\hline
	\textbf{Human Replay Usage} & 971,000 & 309,571 \\
    \hline
    \textbf{Roles in League} & MA, ME, LE & MA, ME, LE, SE, EE, AEE \\
    \hline
    \textbf{Consuming Speed} & 390 fps/TPU core & 210 fps/GPU \\
    \hline
    \textbf{Generating Speed} & 195 fps received by each TPU core & 43 fps received by each GPU \\
    \hline
    \multirow{4}{*}{\textbf{TPU/GPU Usage}} & 3,072 third-generation TPU cores in total with & \\
    & half for training and another half for inference & 144 Nvidia Tesla V100 GPUs (without RDMA) \\
    & (each of the 12 agents uses 128 TPU cores for & with 96 for training and 48 for inference \\
    & training and 128 TPU cores for inference) & \\
    \hline
    \textbf{CPU Usage} & 50,400 preemptible CPU cores & 13,440 standard CPU cores \\
    \hline
    \textbf{Concurrent Matches} & 192,000 & 6,720 \\
    \hline
    \textbf{Total Training Time} & 44 days & 57 days \\
    \hline
    \hline
  \end{tabular}
  }
\end{table*}


\section{Related Work}
Solving challenging games has been considered as a straightforward way to validate the advance 
of artificial intelligence. Our work thus belongs to this series that using general deep 
reinforcement learning to play games. As mentioned previously, the games of 
GO~\cite{silver2016mastering,silver2017mastering}, VizDoom~\cite{kempka2016vizdoom}, 
Quake III Arena Capture the Flag~\cite{jaderberg2019human}, and Dota 2~\cite{openai-five} 
has already been mastered using deep RL, evaluated by expert human players. These games are all 
competitive games that at least two opposite camps of players should join the game to compete. 
This naturally motivates the development of multi-agent learning and game-theoretic algorithms. 
Population-based training, with fictitious self-play as a specific case, thus becomes
one of the most effective way to generate strong AIs. Known as its game complexities on 
both strategic and operational domains, StarCraft has been pushed on the top of AI challenges. 
Prior to AlphaStar, there have been a number of attempts focusing on decomposing SC2 into
various mini-games, such as scenarios of 
micro-management~\cite{usunier2016episodic,sc2le,han2019grid,du2019liir}, 
build order optimization~\cite{elnabarawy2020starcraft}, resources gathering~\cite{sc2le}, etc. 
Our previous work~\cite{sun2018tstarbots} focused on SC2 full game and started
to use RL to train agents based on some high-level actions. Then, AlphaStar~\cite{astar} 
is released, using human-comparable operations and general RL with a population of training 
agents to reach grandmaster level performance. After that, the domain is still active, 
with efforts paid on many aspects of the game, such 
as build order optimization and sub-goal 
selection~\cite{wang2020roma,elnabarawy2020starcraft,xu2020hierarchial}. 
Also, a Terran AI shows in an exhibition to win some matches against professional human 
players~\cite{wang2020scc}, using similar methods as AlphaStar except that it does some 
optimization over AlphaStar's neural network architecture and uses 
three main agents in the league, while a thorough assessment of these 
differences are missing, and the implementation details and many experimental configurations 
remain unclear.

Our work is most related to AlphaStar. We attempt to understand every detail using deep RL and 
league training to play SC2, with orders of less computations compared to AlphaStar. 
We demonstrate the advantages
of the proposed techniques of DLT, RGPS and DAPO with comprehensive studies and comparisons with
AlphaStar Surrogate. We aim to provide an open-sourced 
system that might directly contributes to both academic and industrial communities. 
Generally, using nowadays' learning methods, the resources required by mastering a game are 
often proportional to the complexity of that game. We could reference some accurate numbers to 
name a few. A distributed version of AlphaGo used 176 GPUs and 1,202 CPU cores for inference with
MCTS~\cite{silver2016mastering}, OpenAI Five for solving DOTA2 used 1,536 GPUs and 172,800 CPU 
cores for training~\cite{openai-five}, and AlphaStar used 3,072 third-generation TPU cores and 
50,400 preemptible CPU cores for training~\cite{astar}. In our work, we used 144 GPUs and 13,440 
CPU cores on the Tencent Cloud.

\section{Preliminaries}
\label{sec:prelim}

\subsection{The Game of StarCraft II}
\label{sec:sc2}
We understand that not all readers are familiar with SC2, so we start with a brief introduction of its 
basic game concepts. SC2 players can feel free to skip this section. 
In SC2, each player needs to make a series of timely decisions to play against the opponent, 
including resource gathering and allocation, building construction, training units, deploying the army, 
researching technologies, and controlling units, etc. The total number of units needed to be controlled 
at a single step might be up to hundreds, and the maximum of this number is set to 600 in our implementation.
The units of the opponent are hidden to the player, unless they appear in the view of the units controlled by 
the player.
In SC2, there are three races: Zerg, Terran and Protoss, that humans can choose any one of them to play the
game. For Zerg, we illustrate a few units frequently appeared in human replays. An important building is the 
Hatchery, which is the base for resource gathering and it can morph to Lair which can further morph to Hive. With
Hatchery, units related to Queen, Zergling, Baneling, Roach and Ravager can be produced. With Lair, units 
related to Hydralisk, Lurker, Infestor, Swarm Host, Mutalisk, Corruptor and Overseer can be produced. 
With Hive, Viper, Brood Lord and Ultralisk can be produced. Among those, Mutalisk, Corruptor, Overseer, Viper and
Brood Lord are air units, and all other units are ground units. Some of these names might be used in the
rest of this paper. Detailed descriptions of all the Zerg units can be found in the SC2's official website 
or~\cite{liquipedia}.

\subsection{Reinforcement Learning}
\label{sec:rl}
The problem considered by RL can be formulated by a tuple $<\AA,\SS,\OO,\PP,r,\gamma,\pi>$. 
At each time step $t$, $s_t\in\SS$ is the global state in the state space 
$\SS$; for partially observed environment, like SC2, we can not access $s_t$ and instead we observe $o_t\in\OO$;
$\AA$ is the action space and $a_t\in\AA$ is the action executed by the agent at time step $t$; 
the probability $P(s_{t+1}|s_t,a_t)$ is the state transition function; $r(s_t,a_t)$ is the 
reward function; $\gamma\in[0,1]$ is a discount factor. Let $\pi(a_t|o_t)$ denote a stochastic policy conditioning
on partial observation $o_t$,
and let $\eta(\pi)=\mathbb{E}_{s_0,a_0,\cdots}\left[R_0\right]$ with $R_t=\sum_{l=0}^{\infty}\gamma^l r_{t+l}$ 
denoting the expected discounted reward, where $a_t\sim\pi(a_t|o_t)$ and $s_{t+1}\sim P(s_{t+1}|s_t,a_t)$. 
At time step $t$, the value function is defined as $V_\pi(s_t)=\mathbb{E}_{a_{t},s_{t+1},\cdots}\left[R_t\right]$. 
For competitive multi-agent RL, centralized value function is often utilized during the training phase to provide
more accurate policy evaluation. That is, for learning the value function, the global state is assumed known.
Solving the RL problem requires to find the optimal policy $\pi^*$ that achieves the maximum expected 
reward $\eta(\pi^*)$.

Actor-critic (AC) algorithm is one of the most commonly adopted RL methods. It uses a parameterized policy 
$\pi_{\theta}$ and update its parameters by directly maximizing the expected reward 
$J(\theta)=\mathbb{E}_{s,a}\left[R\right]$ using the policy gradient
\begin{align}
\nabla_{\theta}J(\theta)=\mathbb{E}_{s,a}\left[\nabla_{\theta}\log\pi_{\theta}(a|s)A_{\pi}(s,a)\right],
\nonumber
\end{align}
where $A(s,a)$ is the critic. There exists several ways to estimate $A(s,a)$, resulting in a variant of AC 
algorithms, including some state-of-the-art representatives like TRPO~\cite{schulman2015trust}, 
PPO~\cite{schulman2017proximal}, V-trace~\cite{espeholt2018impala} and UPGO~\cite{astar}. In this paper, we
will use a combination of the PPO, V-trace and UPGO algorithms.

\subsection{Self-Play, Population-based Training and Populating the League}
\label{sec:seflplay}
Self-play is a training scheme and it arises in the context of multi-agent learning,
where a learning agent is 
trained to play against itself and/or its historical checkpoints during training.
This method is game-theoretically justified,
as it turns out to be a Monte Carlo implementation (i.e., sampling the opponents) of the so-called fictitious 
self-play to approximate the Nash Equilibrium~\cite{lanctot2017unified,balduzzi2019open,srinivasan2018actor}. 
Once an opponent is sampled, it can be considered as part of the environment's dynamics from the perspective
of the learning agent, so that single agent RL algorithms mentioned in Section~\ref{sec:rl} can be utilized, 
serving as a proxy algorithm to learn the best response to its opponent~\cite{lanctot2017unified}.
For example, AlphaGo~\cite{silver2016mastering} is trained by self-play with MCTS tree search. 
Based on the concept of self-play, population-based training (PBT) has been proposed. In PBT, 
agents are trained independently with different objectives that are learned or shaped dynamically. 
PBT is shown to be successful in the first-person video game, Quake III Arena Capture the 
Flag~\cite{jaderberg2019human}. 
Similar to PBT, AlphaStar introduced populating the league~\cite{astar}, an approach to maintain a league of
growing number of diverse training agents that can play against each other.

\subsection{Distributed Learner-Actor Architecture}
\label{sec:actor-critic-arch}
There have been continuous attempts to scale up RL by distributed systems.
For example, A3C~\cite{mnih2016asynchronous} proposes a framework with distributed independent actors.
Each actor has its own copy of the policy and the gradients are asynchronously aggregated on a single 
parameter server. GA3C~\cite{babaeizadeh2016reinforcement} extends A3C by using a GPU for both inference 
and policy learning in an asynchronous mode.
IMPALA~\cite{espeholt2018impala} proposes a learner-actor architecture, in which distributed actors are 
responsible for data generation and the learner receives the data for gradient computation. 
SEED RL~\cite{espeholt2019seed} improves upon the IMPALA architecture and achieves higher throughput in 
both single-machine and multi-node scenarios by using TPUs for inference on the learner. 
In this paper, we will use a distributed learner-actor architecture named TLeague~\cite{sun2020tleague} 
that is similar to SEED RL, while our
architecture is specifically designed to solve multi-agent learning problems.


\section{Technical Details and Methods}
\subsection{Environmental Engineering}
To push the frontier of AI research, Google's DeepMind and Blizzard Entertainment jointly created 
SC2LE~\cite{sc2le}, providing APIs to allow outer control of the game. DeepMind further wraps the core library 
in Python as PySC2. Our work is based on an early version~of PySC2, with extensions to expose more
interfaces to fetch information like unit attributes, game status, technology tree, and version control, etc. 
The exposed information is necessary to let the AI agent better understand the game status, and the information 
can also be acquired by human players in the game. These interfaces are implemented in 
a convenient way through the general environmental framework Arena \cite{wang2019arena}.

\subsection{Observation, Action and Reward}
\label{sec:space}
Similar to AlphaStar, the spaces within which the agent can observe and act are comparable to humans. That is,
at each step, the observation of the agent consists of the attributes of the visible units, a spatial 
2D feature, a vector of global statistics, the last action the agent executed, and some masks encoding
the game logic (e.g., after choosing a Zergling, `Build' abilities are not available), 
without any information hidden in fog. 
Similar to humans, the agent acts by choosing an ability, selecting executors and targets, 
and suffering from delays. 
Nevertheless, we have to admit that we indeed remove the camera constraint like the demonstration 
version of AlphaStar to facilitate the learning. The APM/EPM reported in Table~\ref{tab:com} shows that the 
agent does not benefit much from its action space, and TStarBot-X tends to operate 10\%/7\%
slower than AlphaStar in average APM/EPM and 49\%/37\% slower in terms of max APM/EPM. 
Figures~\ref{tab:obs} and~\ref{tab:act} show the detailed observation and action spaces.

\begin{table*}[!ht]
  \centering
  \small
  \caption{Observation space in TStarBot-X. In the following table, UC is short for `Unit Count'; stat. indicates statistics; B \& R prog. is short for `Building and Researching progress'. BO is abbreviation for `Build Order'; Cmd indicates `Command'; NOOP means `No-Operation'.}
  \label{tab:obs}
  \scalebox{0.76}{
  \begin{tabular}{c|c|c|c|c|c|c|c}
    \hline
    \hline    
    Obs: frames & Unit attributes & Unit coordinate & Unit buff & Unit order & Unit type & Images & Player stat. \\
    \hline
    Space & (600, 369) & (600, 2) & (600,) & (600,) & (600,) & (128, 128, 15) & (10,) \\
    \hline
    \hline
    Obs: frames & Upgrade & Game length & Self UC & Enemy UC & Anti-air UC & B \& R prog. \\
    \hline
    Space & (28,) & (64,) & (57,) & (57,) & (4,) & (92,) \\
    \hline
    \hline
    Obs: $z$-stat & BO & BO coordinate & Unit stat. \\
    \hline
    Space & (20, 62) & (20, 18) & (80,) \\
    \hline
    \hline
    Obs: last action & Ability & NOOP & Queued & Selection & Cmd unit & Cmd position \\
    \hline
    Space & (117,) & (128,) & (2,) & (64, 601) & (600,) & (128, 128) \\
    \hline
    \hline
    Obs: mask & Ability & Selection & Cmd unit & Cmd position \\
    \hline
    Space & (117,) & (117, 600) & (117, 600) & (117, 128, 128) \\
    \hline
    \hline
  \end{tabular}
  }
\end{table*}

\begin{table*}[!ht]
  \centering
  \small
  \caption{Action space in TStarBot-X. NOOP indicates `No-Operation' and it is the same as the `Delay' head used in AlphaStar. For the selection head, we need one additional label to indicate the selection termination, which is similar to the end of sequence (EOS) in natural language processing.}
  \label{tab:act}
  \scalebox{0.8}{
  \begin{tabular}{c|c|c|c|c|c|c}
    \hline
    \hline
    Action heads & Ability & NOOP & Queued & Selection & Cmd unit & Cmd position \\
    \hline
    Space & (117,) & (128,) & (2,) & (64, 601) & (600,) & (128, 128) \\
    \hline
    \hline
  \end{tabular}
  }
\end{table*}

In Table~\ref{tab:obs}, $z$-stat is a high-level feature introduced and highlighted in AlphaStar.
By sampling a human replay, a $z$-stat is fixed that it includes a build order sequence of length 20, 
the coordinate of each unit in that order and a vector indicating the unit appearance in the sampled replay.
It has been emphasized in~\cite{astar} that the strategic diversity of the policy can be controlled by varying the $z$-stat.
In Table~\ref{tab:act}, the no-operation (NOOP) head allows a maximum delay of 128 game frames, 
around 5.7 seconds in real-time game. The selection head can select a maximum number of 64 units at a step.
The maximum number of observed units is set to 600, a number determined by parsing all human replays that only 
less than 2\% frames of human replays contain more than 600 units.
A detailed list of used features and abilities can be found via the open-sourced repository.
For reward function, we follow AlphaStar's configuration by using the win-loss outcome, the Edit distance between 
the build order in $z$-stat and the immediate build order, and the Manhattan distances between the build units, 
effects, upgrades in $z$-stat and the according immediate values. Except the win-loss reward, each of the other
rewards is activated with a probability of 0.25 for the main agent. Exploiters initialized from the standard supervised agent 
only use the win-loss reward. Exploiters initialized from diversely fine-tuned supervised agents use all the above
rewards with 25\% activation probability on the build order reward and 100\% activation for others. 
The discount factor $\gamma$ is set to 1.0, i.e., without discounts.

\subsection{Neural Network Architechture}

\begin{figure*}[t]
\center
\includegraphics[width=1.0\linewidth]{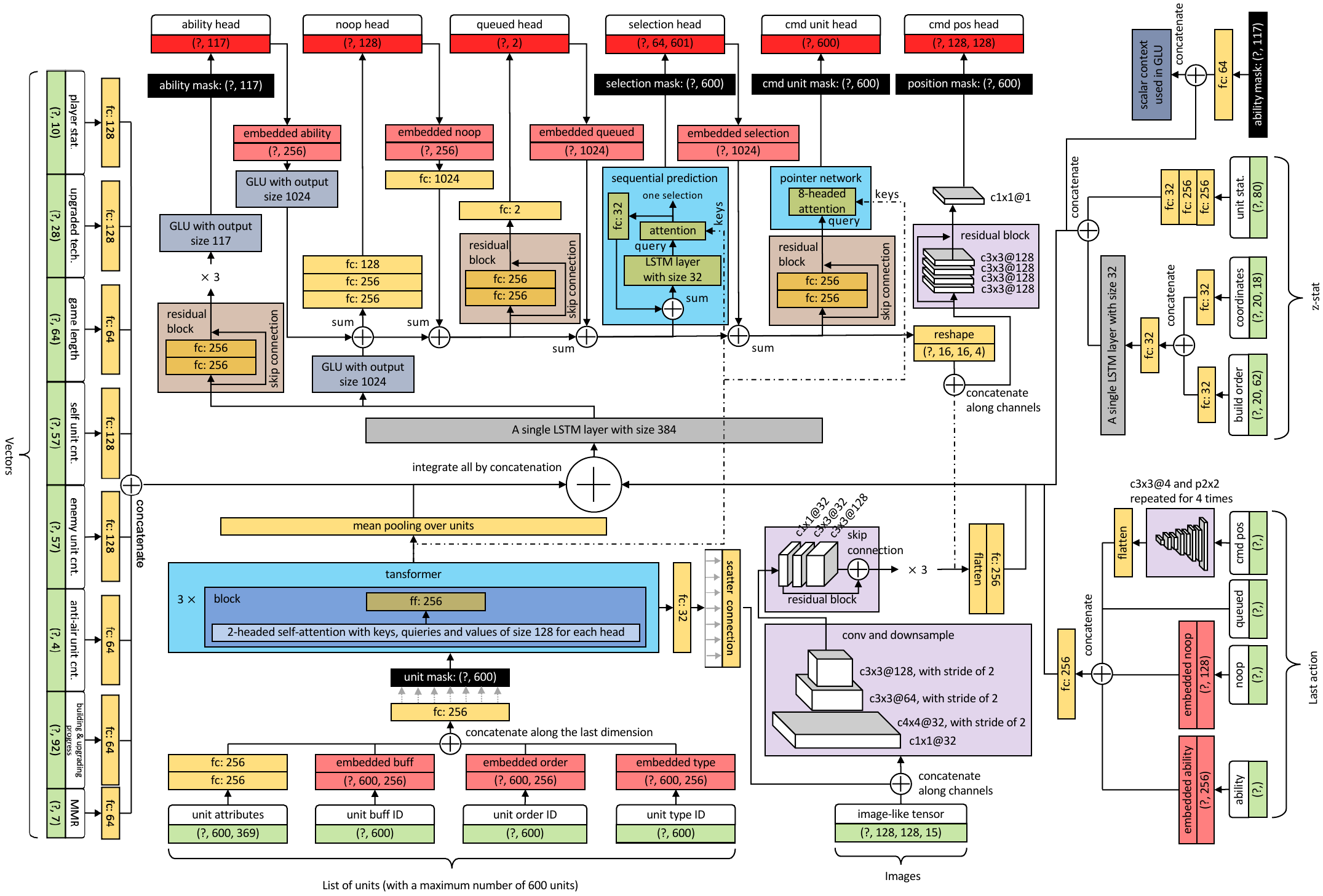}
\caption{The detailed policy architecture. `?' indicates the batch dimension. All green rectangles are inputs. Dark red rectangles are output multi-head actions. Black rectangles denote masks relying on the game logic.}
\label{fig:pi_arc}
\end{figure*}

\begin{figure*}[t]
\center
\includegraphics[width=1.0\linewidth]{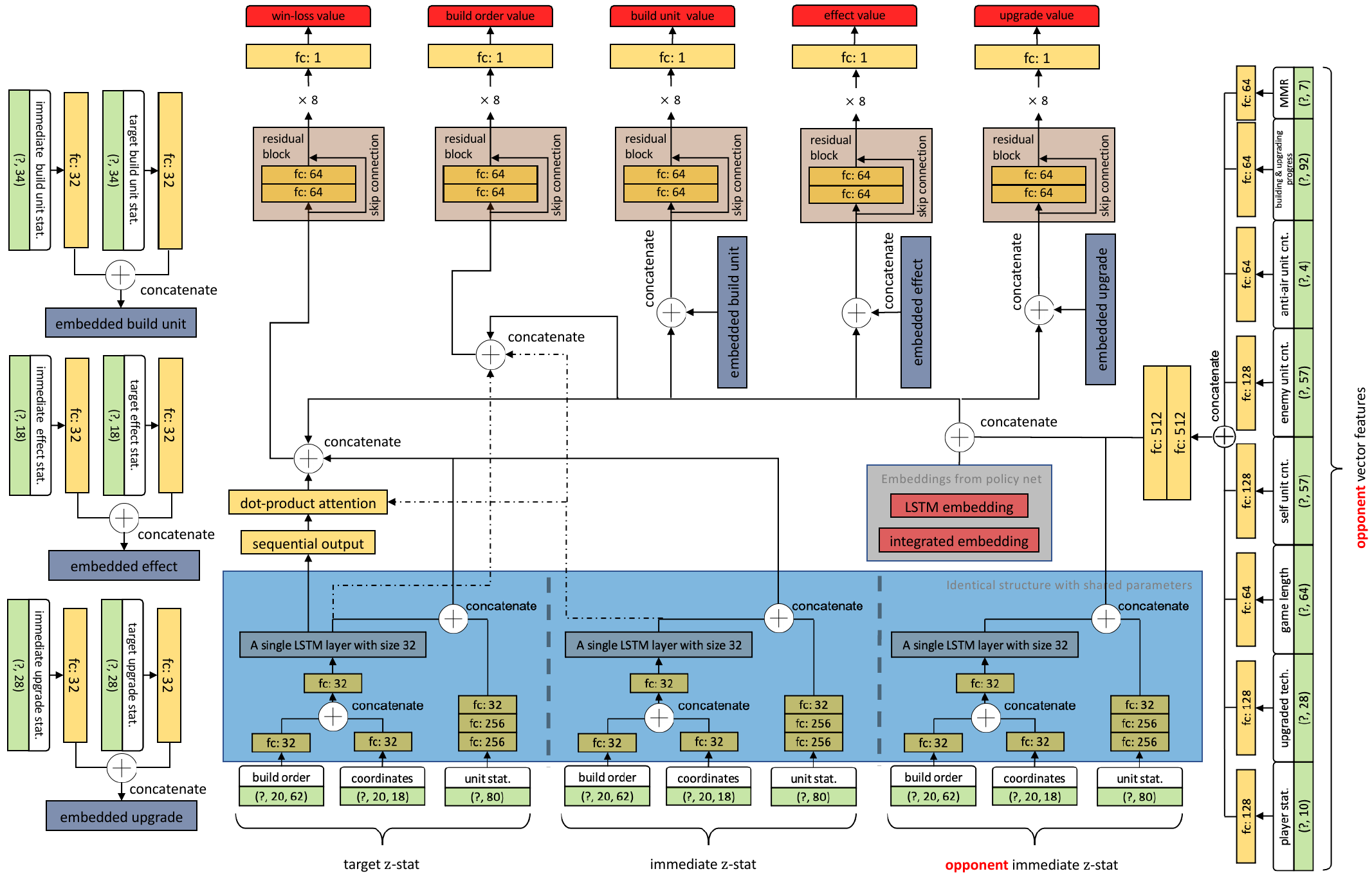}
\caption{The detailed value architecture. `?' indicates the batch dimension. All green rectangles are inputs. Dark red rectangles are output multi-head values. The light red rectangles, i.e., LSTM embedding and integrated embedding are the embeddings just after and before the central LSTM layer in the policy net.}
\label{fig:v_arc}
\end{figure*}

Figures~\ref{fig:pi_arc} and~\ref{fig:v_arc} show the detailed policy and value net architectures used by TStarBot-X, respectively. 
Most of the functional modules, such as a central LSTM 
for memorization, a transformer for encoding list of unit features, sequential prediction for the selection head, and 
pointer network for choosing a target unit, etc., follow what are used in AlphaStar. The main differences are: 
\begin{itemize}
	\item we significantly reduce the size of the network by removing some functional modules and reducing the number of dense layers and layer size. For example, we remove the FiLM~\cite{astar} connections; we only use GLU for the ability head; we use a single layer for the central LSTM; for residual blocks, which are adopted multiple times, we never stack the blocks more than 3 times in the policy net. By doing so, both the training and inference speed becomes endurable under our computation scale;
	\item we replace the transformer connections for build order embedding with a single LSTM layer, since compared to the units embedding, build order is sequentially mattered. We indeed experiment with ablations and the results show that a single LSTM layer here is slightly superior than using transformers during imitation learning;
	\item for value net, differing from AlphaStar which creates a symmetric representation using the entire network for both self and opponent's observations, we only fetch some general statistics from the enemy side into a single short vector. As shown in Figure~\ref{fig:v_arc}, it includes the player statistics, upgraded technologies, unit counts, etc. This, together with the opponent's immediate $z$-stat embedding, are the only inputs from the opponent side. The value still acts in a centralized way while it evaluates by only conditioning on the enemy's strategic statistics and self information. By doing so, the value net is around 30 times smaller than that used in AlphaStar.
\end{itemize}
It is also worth mentioning a number of attempts before choosing a condensed version of AlphaStar's network. Prior 
to the release of AlphaStar's technical details, we indeed tried multiple versions of neural network compositions:
\begin{itemize}
	\item for the selection head, we first tried multi-binary output that each unit is independently determined to be selected or not at the last layer, while they share some common units representation at lower layers. Multi-binary head generally works well and is much more efficient compared to sequential predictions. However, it unavoidably suffers from the unit independency that a few units intended to be selected are often missed from the prediction with probability;
	\item we parallelly investigated the 	`scatter connection' which was then demonstrated to be effective in~\cite{astar}. This operation scatters the unit embeddings to the corresponding spatial positions in the 2D feature representation by concatenation along the channels and appending zeros for other locations without any units filled. It is similar to the feature representation used in GridNet~\cite{han2019grid}. Moreover, we tried `gather connection', an affine and inverse operator of scatter connection that fetches the output of the convolution layers according to the unit positions and append the fetched dense representations to the unit embeddings. By using scatter connection and gather connection together, we initially abandoned transformers and the units' interaction was completely captured by convolution layers. This is still for efficiency consideration to remove the $O(n^2)$ complexity in transformers (with $n$ being the number of units). It turns out that the trained agent prefers to control a group of units that are spatially close to each other and barely controls one or a few units for specific micro-management;
	\item absorbing the selection operation into abilities is a fairer way to make an AI agent act like human. In Figure~\ref{fig:pi_arc} and AlphaStar, selection is considered as an independent action head as a part of the auto-regressive structure. This naturally gives an AI agent superhuman capability that it can always finish selection and an effective command, e.g., attack a target, at a single step. On the contrary, humans have to perform a selection operation at an earlier step and then launch the command after that. We empirically find that by isolating selection operation from other abilities into the auto-regressive structure, the AI agent can always reach the ratio of EPM/APM around $0.8\sim0.85$. An earlier version of TStarBot-X indeed absorbs the selection head into the ability head, and it has to use a single step to act selection first and then executes a command at a later step, identical to human. The supervised agent can then only reach an average ratio of EPM/APM around 0.5.  
\end{itemize}

\subsection{Imitation Learning with Importance Sampling}
\label{sec:IL}
We use a dataset containing 309,571 Zerg vs. Zerg replays. Table~\ref{tab:rep} shows the replay distribution 
with respect to game versions. Version 4.8.3 seems to be damaged and it raises errors when loading replays in this version. 
\begin{table*}[!ht]
  \centering
  \small
  \caption{Human replays used by TStarBot-X.}
  \label{tab:rep}
  \begin{tabular}{c|c|c|c|c|c|c|c|c}
    \hline
    Game version & 4.8.2 & 4.8.4 & 4.8.6 & 4.9.0 & 4.9.1 & 4.9.2 & 4.9.3 & total \\
    \hline
    Number of replays & 43,144 & 31,576 & 51,907 & 33,612 & 29,467 & 44,152 & 75,713 & 309,571 \\
    \hline
  \end{tabular}
\end{table*}
The replays log the human players' raw actions and generally each one can be mapped into a valid action
under the space introduced in \ref{sec:space}. However, considerable amount of data preprocessing efforts 
are required. For example, `Smart' over a `Hatchery' is equivalent to the action `Rally', while `Smart'
attackable units on an enemy unit indicates `Attack' that unit. That is, raw actions from replays labeled
with the same abilities work differently when the executors are from different types. To remove ambiguity, 
we compact the raw actions and a list of the preprocessed abilities for Zerg can be found in the open-sourced
repository via the provided link.

In~\cite{astar}, there are very few discussions on imitation/supervised learning, while we empirically find 
that a straightforward run over the preprocessed data is not a good choice. Importance sampling over abilities is
extremely important. Using a simple importance sampling configuration will increase the win-rate of the agent 
against the Elite-bot from 68\% to 85\% (90\% win-rate mentioned in introduction is achieved by further fine-tuning
the model on data with higher MMR). That is, we downsample 	`NOOP' by a rate of 0.2 and downsample `Smart' 
by a rate of 0.25, while all other abilities that fail to appear in each replay at least once in average will be
upsampled by the ratio the number of total replays over the ability's total count, capped by 10.0. The actions
that do not belong to the above mentioned cases remain their original distribution. Since we
use LSTM, importance sampling should be implemented over trajectories. Specifically, we sample trajectories
by summing up all the point-wise sampling weights within a trajectory, normalized by the total weights in the
concurrent replay memory. After a trajectory is sampled, the point-wise weights are reassigned to the supervised 
loss for back-propagation. This guarantees the estimation unbiased.

Our imitation learning procedure differs from traditional supervised learning fashions where the 
data is preprocessed once and then stored untouched. Instead, we use online and distributed data generators, 
which are named Replay Actors, to send human trajectories to the imitation learner as the training proceeds.
Although such an infrastructure uses more CPUs to generate data, it keeps compatible with the RL infrastructure and
facilitate algorithms using mixture of human data and self-generated data. Another benefit is that feature
engineering is much more convenient. The training is implemented on 144 GPUs and 5,600 CPU cores.
We choose a model after 43 hours of imitation learning, which shows the best testing performance against the 
Elite-bot, and then we fine-tune this model on 1500 human replays with at least one human player's MMR 
above 6200. The obtained model, with a win-rate of 90\% against the Elite-bot, servers as a standard baseline 
supervised agent, which will be used as one of the initial models for exploiters in league training.

From the baseline supervised agent, we continue fine-tuning this model in $K=6$ separate subsets of human replays
with at least one human player's MMR above 3500 (the subset of replays may overlap).
Each subset of the replays covers the usage of a specific unit or strategy, including the Hydralisk, Lurker,
Mutalisk, Nydus Network, Researching TunnelingClaws and Worker Rush, which are
frequently observed in human replays. The $K$ fine-tuned supervised models will be used by some specific exploiters
as initial models, as will be introduced in \ref{sec:DLT}. 

\subsection{Diversified League Training}
\label{sec:DLT}
Choosing appropriate RL algorithms, e.g., V-trace, PPO or UPGO, could stabilise RL training. We have empirically 
verified that using prioritized fictitious self-play (PFSP)~\cite{astar}, 
optimizing Equation~\ref{eq:loss} or replacing V-trace with PPO in Equation~\ref{eq:loss} can both let the 
training agent consistently surpass its historical models. However, due to the complexities
in SC2, especially for its large space of cyclic and non-transitive strategies, using PFSP is not sufficient to let
the agent discover robust or novel policies. Diversity in the policy population thus plays an important role to
solve this challenge.

To enrich the diversity in the policy population, AlphaStar emphasizes the usage of the $z$-stat feature and populating
the league~\cite{astar}. The former one directly fits strong strategic human knowledge into the policy, enabling the 
control of the agents' strategies by just choosing the $z$-stat, while the latter one focuses on game-theoretic perspective
to employ exploiters to discover new strategies. As we have explained, the former one aims to enhance the diversity in MA 
(in AlphaStar, only MAs use $z$-stat), and the latter one enriches the exploiters' diversity in the league.
To be general, we refer to populating the league with diverse 
roles as League Training. The word `League' is essentially from Blizzard Entertainment's Battle.net League. 
AlphaStar's league
contains three agent roles: Main Agent (MA), Main Exploiter (ME) and League Exploiter (LE). MA is unique in the league (for a distinct race)
and it keeps training against all other agents and itself throughout the entire life of learning. 
It is positioned as a ROBUST agent that can respond to all other strategies. ME is set to exploit MA only and LE aims to
find weakness of the entire league. Both ME and LE reset to the supervised model periodically when their 
maximum number of training steps is reached or they achieve a high win-rate (above 70\%) against their opponents. 
It can be deduced from~\cite{astar} (or Table~\ref{tab:com}) that the ME of AlphaStar resets every 24 hours approximately ($4\times10^9$ steps) and the LE 
resets about every 2 days in average ($2\times10^9$ steps with a reset probability of 25\%), regardless of their win-rates.
It is also worth noting that from Figure 4A in~\cite{astar}, the performance of LE keeps improving even if it resets
every 2 days in average. This implies that restarting from a baseline supervised model, AlphaStar's exploiters
can quickly reach a strong performance that is competitive to its MA (which never reset) 
even within a single training period. In addition, a mixture training across all 
SC2 races further diversifies the league.

In our experiments, reimplementing the above configured league training raises issues: 
1) keeping the same amount of periodic training steps for the exploiters costs us more than 10 days for a single period. 
This is unendurable and makes the league too sparse; 2) using less training steps leads to 
weak ME and LE agents that can hardly catch up with the continuously trained MA, and the
problem becomes worse as the training proceeds; 3) Zerg vs. Zerg reduces the chance for discovering 
many novel strategies, and strategic explorations become more difficult.

To address the aforementioned issues, we introduce several new agent roles:
\begin{itemize}
	\item Specific Exploiter (SE): it works similarly as ME except that it starts from one of the specific fine-tuned supervised model introduced at the end of~\ref{sec:IL}. Each SE focuses on the exploration of a specific strategy or unit type. Although SE is limited by its distinct strategy or unit type, it still resets like ME to have chance to discover novel timing and micro-management policies. SE only uses the win-loss reward and samples $z$-stats from the according subset of human replays. It utilizes policy distillation to stay close to its specific initial policy with a larger coefficient compared to others.
	\item Evolutionary Exploiter (EE): it plays against MA and periodically resets to (inherits) a historical EE with the best win-rate against MA. Once a maximum number of steps reached or it beats MA above 70\% win-rate, a copy of the model is frozen into the league and it resets (inherits). EE focuses on continually exploiting MA probably with some distinct strategies. As can be imagined, EE is easy to over-exploit MA, and hence its learning period length should be carefully set; otherwise, if the learning period is too short, multiple strong EEs with similar strategy will be added into the league frequently, and this will lead to a sharp change in MA's value, as we have observed in our trials. EE's evolutionary path is less likely to be wide but tends to be deep.
	\item Adaptive Evolutionary Exploiter (AEE): it plays against MA and periodically resets to (inherits) a historical leaf AEE node, whose win-rate against MA lies in 20\%-50\%, implying this model is slightly weaker than MA but still exploitable. A leaf AEE node indicates the newest model in a branch which has never been inherited yet. Each branch might keep inheriting and improving some distinct strategy as it grows. If there is no historical model whose win-rate lies in 20\%-50\%, it resets to its initial supervised model. If there exists more than one eligible model, it resets to the one whose win-rate against MA is closest to 50\%. AEE continually adds comparable opponents for MA to the league, making the training of MA stable, and it balances exploration and exploitation. AEE's evolutionary path is less likely to be deep but tends to be wide.
\end{itemize}

\begin{figure}[t]
\center
\subfigure[Evolutionary Exploiter]{
\includegraphics[width=0.45\linewidth]{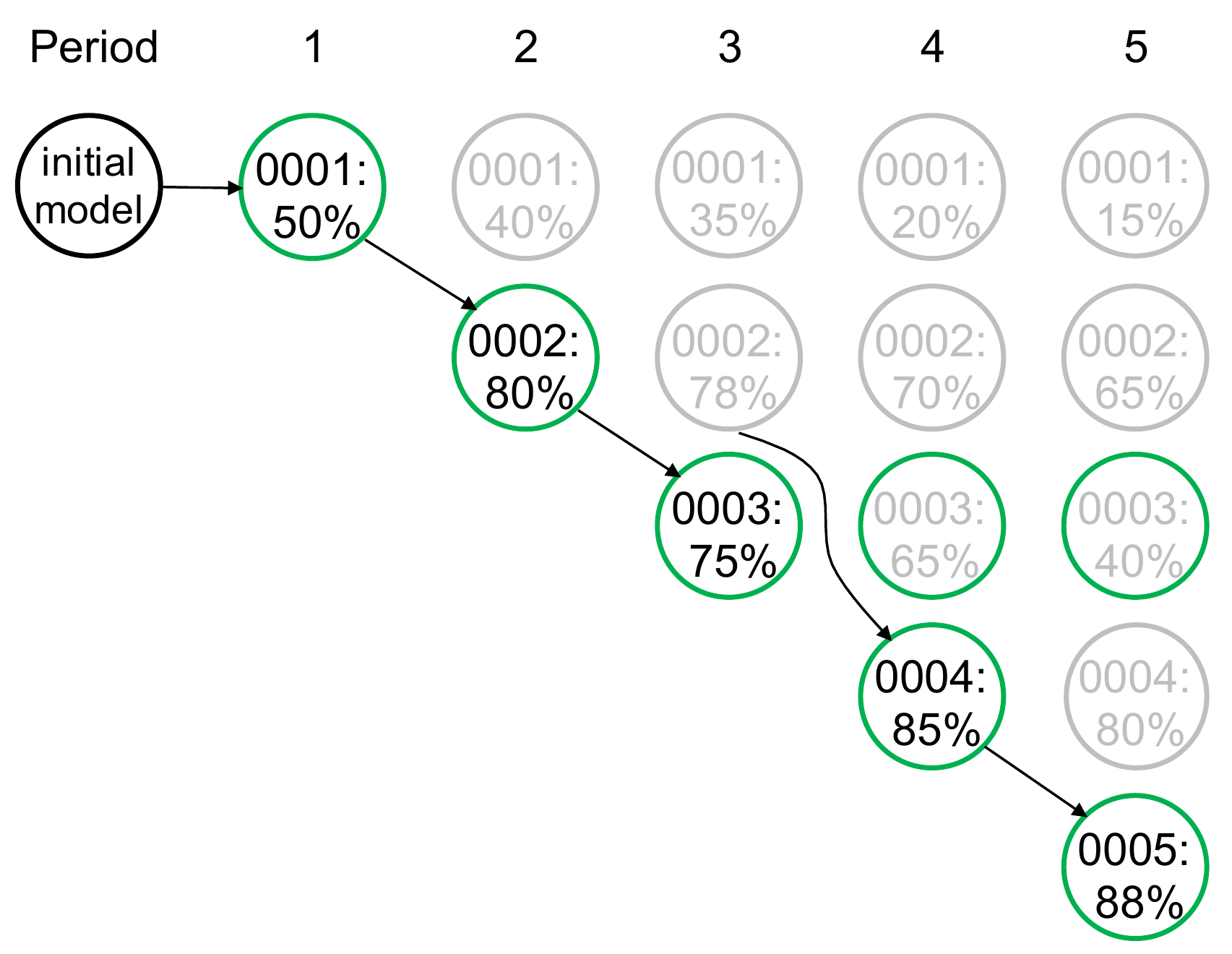}
\label{fig:roles-ee}
}\hskip 0.3in
\subfigure[Adaptive Evolutionary Exploiter]{
\includegraphics[width=0.45\linewidth]{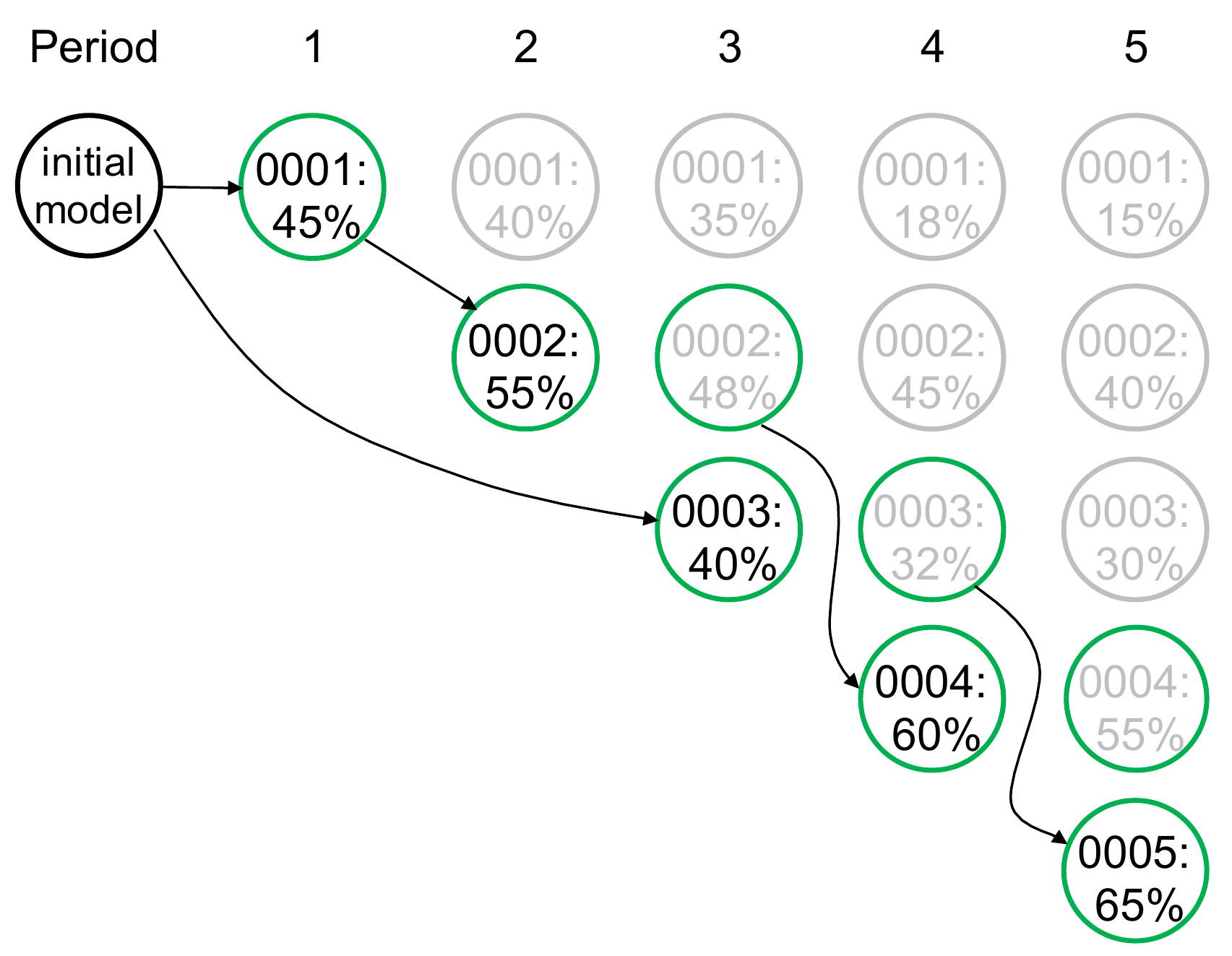}
\label{fig:roles-aee}
}
\caption{Examples of how EE and AEE work as the training periods proceed. Green cycles indicate leaf nodes. The texts in each node consist of a model ID before the colon followed by the win-rate against the current MA at the end of that training period. Grey texts indicate the corresponding node has been frozen as a historical model in the league.}
\label{fig:roles}
\end{figure}

Figure~\ref{fig:roles} shows two examples of how EE and AEE work. We give a brief explanation for 
Figure~\ref{fig:roles-aee}, and then Figure~\ref{fig:roles-ee} can be understood easily. 
In Figure~\ref{fig:roles-aee}, 
the training period proceeds horizontally and a new model ID is generated for this AEE in a new learning period. 
For a distinct model ID, once it has been fixed and added into the league, 
its win-rate against MA will generally decrease 
horizontally, since MA keeps training with probability to sample it as opponent. 
At the beginning of period 2, model 0001 is a leaf node and its 
win-rate against MA lies in 20\%-50\%, so model 0002 inherits model 0001; at the beginning of period 3, model 
0002's win-rate exceeds 50\%
and there does not exist any other leaf node, so model 0003 resets to the initial model; 
at the beginning of period 4,
there exist two leaf nodes, i.e., model 0002 with 48\% win-rate and model 0003 with 40\% win-rate, 
and then model 0004 inherits model 0002 whose win-rate is closest to 50\%, to name a few. Both EE and AEE can be combined 
with SE, ME, LE by varying the initial supervised model or the opponent matching mechanism.

The proposed exploiters either reduce the exploration space, extend the training life or seek a balance for stabilising 
MA's training, and hence they are general methods to 
deal with the aforementioned league training problems. 
After some practical trails, in our formal experiment, we use a combination among SE, ME, LE and AEE in the league.
We refer to such league training with a mixture of these exploiters as Diversified League Training (DLT).
We will show that DLT enriches the league diversity and benefits MA's training, compared to AlphaStar Surrogate.
The MA will use 25\% Self-Play, 60\% PFSP and 15\% of PFSP matches against forgotten main players,
similar to AlphaStar. In our experiments, MA is assigned with 32 GPUs and each exploiter is assigned with 8 GPUs for training.
MA pushes a copy of its model into the league every $3\times10^8$ steps (around 12 hours in our case). 
The period length of all the exploiters is set with a minimum number of $7.65\times10^7$ steps (also around 12 hours), and after that the
exploiters periodically check whether a maximum number of $1.53\times10^8$ steps is reached or it satisfies its 
condition to reset, inherit or continue training.

\subsection{Rule-Guided Policy Search}
Following AlphaStar, our RL loss consists of a V-trace~\cite{espeholt2018impala} loss, an UPGO~\cite{astar} loss, 
an entropy regularizer, and a policy distillation~\cite{rusu2015policy} term targeting to the supervised model.
That is, the considered RL loss is
\begin{equation}
\mathcal{L}_{RL}=\mathcal{L}_{V\text{-}trace}+\mathcal{L}_{UPGO}+\mathcal{L}_{entropy}+\mathcal{L}_{distill}.
\label{eq:loss}
\end{equation}
We have also tried a distributed version of PPO~\cite{schulman2017proximal}, and the results suggest
that distributed PPO + UPGO does not show much difference compared with V-trace + UPGO within 5 days of training. 

In addition,
we introduce a new Rule-Guided Policy Search (RGPS) loss to promote MA's exploration on some critical states.
RGPS is a principled way to fuse expert knowledge into a policy.
Specifically,
the RGPS loss induces the policy as follows:
\begin{equation}
\mathcal{L}_{RGPS}=
\begin{cases}
KL\Big(\pi_{expert}(\cdot|s)||\pi_{\theta}(\cdot|s)\Big)\ , & \ s\in\mathcal{S}_{critical}, \\
0, & s\notin\mathcal{S}_{critical},
\end{cases}
\label{eq:rgps}
\end{equation}
where $\pi_{\theta}$ is the training policy parameterized by $\theta$, 
$\pi_{expert}$ is an expert teacher policy 
and $\mathcal{S}_{critical}\subset\mathcal{S}$ is a small subset of state space which are considered to be critical by human experts.
The RGPS loss is activated only when $s \in \mathcal{S}_{critical}$.
RGPS promotes explorations on these states by following the teacher expert policy $\pi_{expert}$ and then generalizes to similar states.
Indeed, 
in SC2 it is extremely difficult for the traditional RL algorithms to explore such critical decisions with limited trial-and-errors.
Still think about the example in the introduction, 
``humans understand that `Hydralisk' is strong against `Mutalisk'...''.
Fitting such hard game logics into a policy requires extremely 
large amount of game episodes, 
because within one match these key state-action pairs, 
say, morphing `Lair' after observing enemy's
intention to use `Mutalisk', 
would probably only be performed once among thousands of steps.
We observe in the experiments that without using RGPS, 
the agents have to spend a large amount of training time to explore some quite simple scenarios, 
such as training `Hydralisk' first requires `Morph Lair'.

For a given $s \in \mathcal{S}_{critical}$ and discrete action space, 
$\pi_{expert}(\cdot|s)$ in Equation~\ref{eq:rgps} can be designated by human expert.
For example, it can be either a one-hot vector as in conventional supervised learning or a soft label via the 
label-smoothing technique~\cite{szegedy2016rethinking}.
We can further use hand-craft rules (by human experts) to decide the critical states.
Conceptually, RGPS is similar to the Guided Policy Search (GPS)~\cite{levine2013guided} in regards of imitating 
an expert policy. The difference is that RGPS is an unified approach for RL by involving an extra 
regularization term, 
while GPS is a two-stage process by first generating the expert policy and then imitating it.
GPS obtains the expert policy via Differential Dynamic Programming (DDP), while RGPS utilizes hand-craft rules.
In most applications, it is impractical to express an expert policy by solving DDP and using closed-form equations, while it is often possible to transfer expert knowledges into
hand-craft rules conveniently using if-else sentences. For example, behavior tree and finite state machine
have been well studied and demonstrated to be practically useful for decades. 
SAIDA~\cite{saida}, a pure rule-based AI in StarCraft: Brood War, won the 2018 AIIDE tournament.
However, an important drawback for systems based on behavior tree, finite state machine or rule-based systems is that 
they lack the ability to learn. That is, these systems can not evolve automatically.
RGPS thus provides a hybrid learning solution to fuse expert knowledge into a parameterized policy through 
RL's exploration.

In our implementation, the `Morph Lair' example mentioned above is an exact case used in RGPS. 
We code a few other game logics in the same way, and we will list all of these logics in the experimental section.
These hand-craft rules might not be optimal, but they can indeed efficiently promote exploration through RGPS
and the policy will then be optimized by policy gradient. 
RGPS is only activated for MA in our experiments.


\subsection{Stabilized Policy Improvement with DAPO}
Divergence-Augmented Policy Optimization (DAPO) \cite{wang2019divergence} was originally derived for 
stabilized policy optimization from the scope of mirror descent algorithms. In the policy space, the mirror
descent step can be explained as penalized policy gradient with a Bregman divergence term between two successive
policies. Choosing a specific form of the function in the Bregman divergence will recover the KL divergence.
In our league training, we apply DAPO at the agent level that the successive policies in the KL divergence are defined as 
two successive stored models with gap between them as a learning period (around 12 hours). 
Empirically, our objective after activating DAPO is
\begin{equation}
\mathcal{L}_{RL}+\mathcal{L}_{RGPS}+\mathcal{L}_{DAPO},
\label{eq:dapo_loss}
\end{equation}
where 
\begin{equation}
\mathcal{L}_{DAPO}=KL\big(\pi_{\theta_{t-1}}(\cdot|s)||\pi_{\theta_t}(\cdot|s)\big),
\end{equation}
and $t$ is the index of the 
learning period. Heuristically, $\mathcal{L}_{DAPO}$ constrains the training policy to stay close to its previous 
historical fixed model, and hence explorations are within a reasonable space, similar to $\mathcal{L}_{distill}$ which forces
the policy to stay close to human strategies. The difference is that the teacher policy shifts in DAPO as the league training proceeds 
while in $\mathcal{L}_{distill}$ the teacher policy is always fixed as the supervised agent. An important necessity for using
DAPO in our case is that our supervised agent is indeed not strong enough. AlphaStar's supervised agent has been validated to reach 
Diamond level in Battle.net officially, while our supervised agent loses some games when playing with Platinum human players half the time.
We conjecture this is due to the significantly reduced network size and imitating only Zerg v.s. Zerg replays.
Therefore, in AlphaStar, even after reseting the exploiters to the supervised agent, the exploiters can still quickly catch up with the 
MA within a single learning period. Hence, consistently constraining MA to stay close to the supervised agent will be a 
natural choice for AlphaStar. However, in our case, situations become totally different. For exploiters, we have proposed 
new agent rules such as AEE to continually training the exploiters to avoid the embarrassed situation that once reseting they will
never catch up with the MA. For MA, apparently, consistently restricting MA around the supervised agent prevents MA from improving
efficiently. Therefore, DAPO is necessary to promote policy improvement. In our implementation, DAPO is only activated for the 
first 4 minutes of an episode and the agent is free to explore after that. As we will show in the experiments, by doing so,
policy improvement can be significantly promoted.

\subsection{The Entire Infrastructure}
\label{sec:infrastructure}
\begin{figure}[t]
\center
\includegraphics[width=0.5\linewidth]{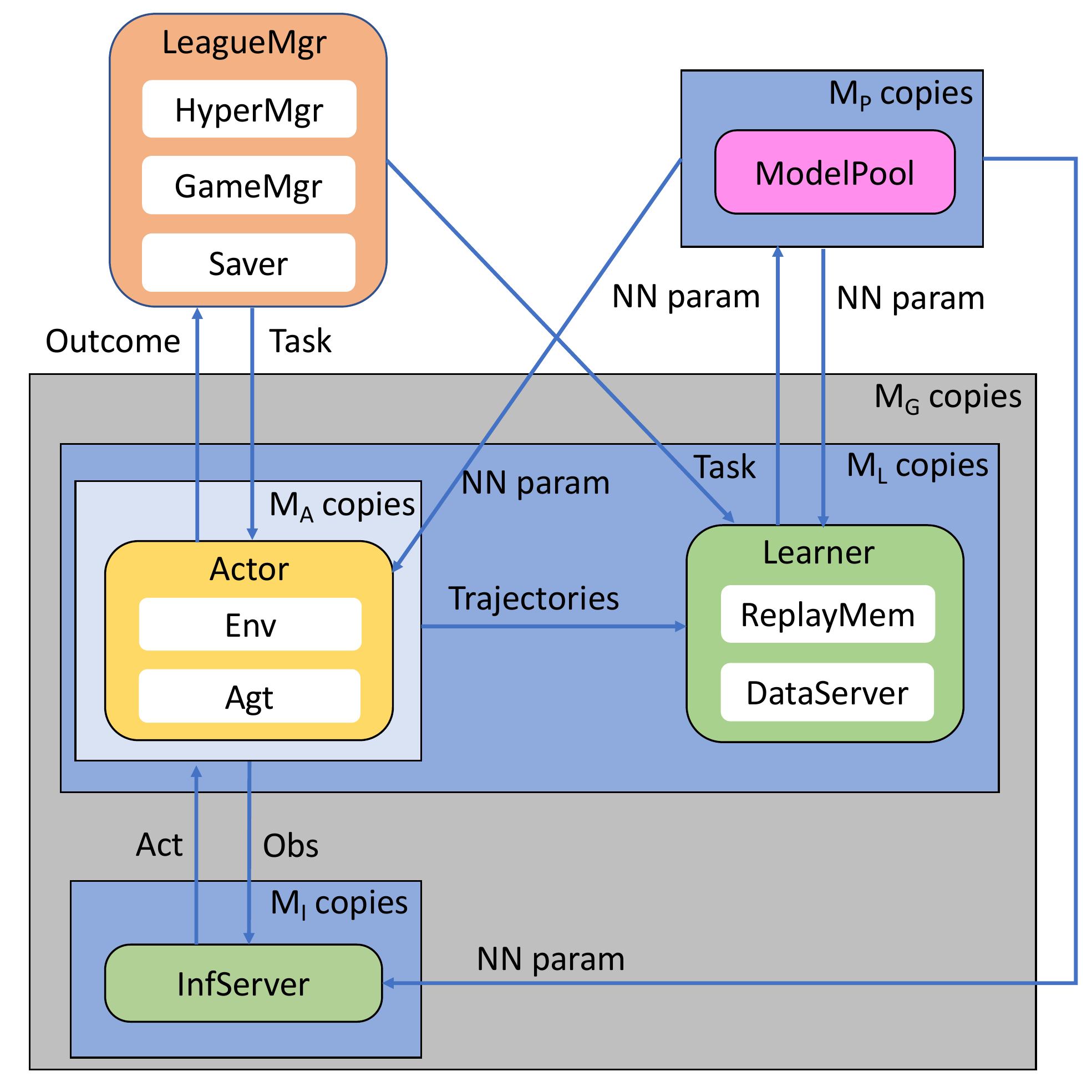}
\caption{
The framework of the entire training architecture TLeague.
}
\label{fig:diagram}
\end{figure}

All the introduced techniques are implemented under an infrastructure depicted in Figure~\ref{fig:diagram}.
To scale up the training,
we adopt a distributed learner-actor architecture,
where the actors are deployed on CPU machines to let the agent interact with the game core to generate trajectories
and send them to the learner,
and the learners are deployed on GPU machines to train the parameters using the received trajectories.
Similar to SEED, we employ a GPU Inference Server (InfServer) to leverage the GPUs for action inference on actors. 
Compared to the scheme of using pure CPUs for action inference, using InfServer 
reduces 34\% of CPU usage while maintaining the same data generation speed.
Importantly, the infrastructure is specially designed for multi-agent learning that a League Manager (LeagueMgr)
schedules the matches and keeps track of all the historical models.
The entire infrastructure is referred to as TLeague~\cite{sun2020tleague}.
The full-scale training is run over a Kubernetes cluster on Tencent Cloud.
Please refer to~\cite{sun2020tleague} for more details. 

\section{Results}
\label{sec:res}
In our experiments, the full scale of computational resources contains 144 NVDIA v100 GPUs and 13,440 CPU cores on Tencent Cloud.
This section includes comprehensive experimental results, including studies of one formal league training experiment 
with full computation scale that lasts for 57 days, several ablation league training experiments each of which 
is running with full computation scale and lasts for 7 days, imitation learning results, and full scale of testing 
matches for evaluation, etc. We access every detail of the learning and show which are the most sensitive parts that affect
the agent performance.

\subsection{Overall Performance}
\subsubsection{Human Evaluation}
Within 57 days of league training of the formal experiment, we invite four human players officially ranked at Master and 
Grandmaster in Battle.net League to evaluate the trained agents. For all the matches, we let MAs play with human players.
We use unconditional policy, i.e., zero $z$-stat, with 50\% probability and randomly sample $z$-stat from the top-25 $z$-stats 
(of the according MA) from the 174 $z$-stats used for training for another half of the time.
There are two modes for humans and AIs to join a game: local and remote. For local connection, the AI agent is deployed on a single 
laptop with a single GeForce GTX 1650 GPU, and the human player uses a laptop under the same local network. 
The agent then suffers an additional 134ms delay (in average, for inference and lag in real time mode) 
except its predicted delay from its NOOP head. For remote connection, the AI agent is deployed on a Tencent Cloud 
Virtual Machine with one NVIDIA v100 GPU, and the human players are free to choose their local equipments. For remote connection,
the agent suffers an additional 179ms delay except its predicted delay from its NOOP head. The invited two Grandmaster players
are at different locations and have to join the game using the remote mode.

The overall human evaluation results are depicted in Table~\ref{tab:res-hum}. Each human player 
is asked to play at least three consecutive matches and additional matches depending on their interests. 
TStarBot-X loses one match against a Grandmaster player. Replays can be downloaded from the public link.
Some evaluations received from the Grandmaster players: ``it can play with diverse strategies across matches'', 
``it locates its Overlords perfectly'', ``its economic operation is professional''. On the other side, the Grandmaster who
won one of five consecutive matches points out that ``it is not good at multitasking and counter-multitasking''.

\begin{table*}[!ht]
  \centering
  \scriptsize
  \caption{Overall results from human evaluation. The score formats as ``Human player : TStarBot-X's MA''.}
  \label{tab:res-hum}
  {
  \begin{tabular}{c|c|c|c|c|c|c}
    \hline
    \hline
    Human player rank & League region & Expertise & Played & Mode/Lag & Score & Training days \\
    \hline
    Master & CN & Protoss & Zerg & Local & 0:11 & 47 \\
    \hline
    Master & CN & Protoss, Terran & Zerg & Local & 0:13 & 49 \\
    \hline
    Grandmaster$^*$ & US & Zerg & Zerg & Remote & 0:3 & 55 \\
    \hline
    Grandmaster & KR, CN (Top-50) & Zerg, Protoss & Zerg & Remote & 1:4 & 55 \\
    \hline
    Grandmaster$^*$ & US & Zerg & Zerg & Remote & 0:4 & 57 \\
    \hline
    \hline
    \multicolumn{7}{l}{\scriptsize * The same human player played on different dates.} \\
  \end{tabular}
  }
\end{table*}


\subsubsection{League Evaluation}

\begin{figure}[t]
\center
\subfigure[]{
\includegraphics[width=0.45\linewidth]{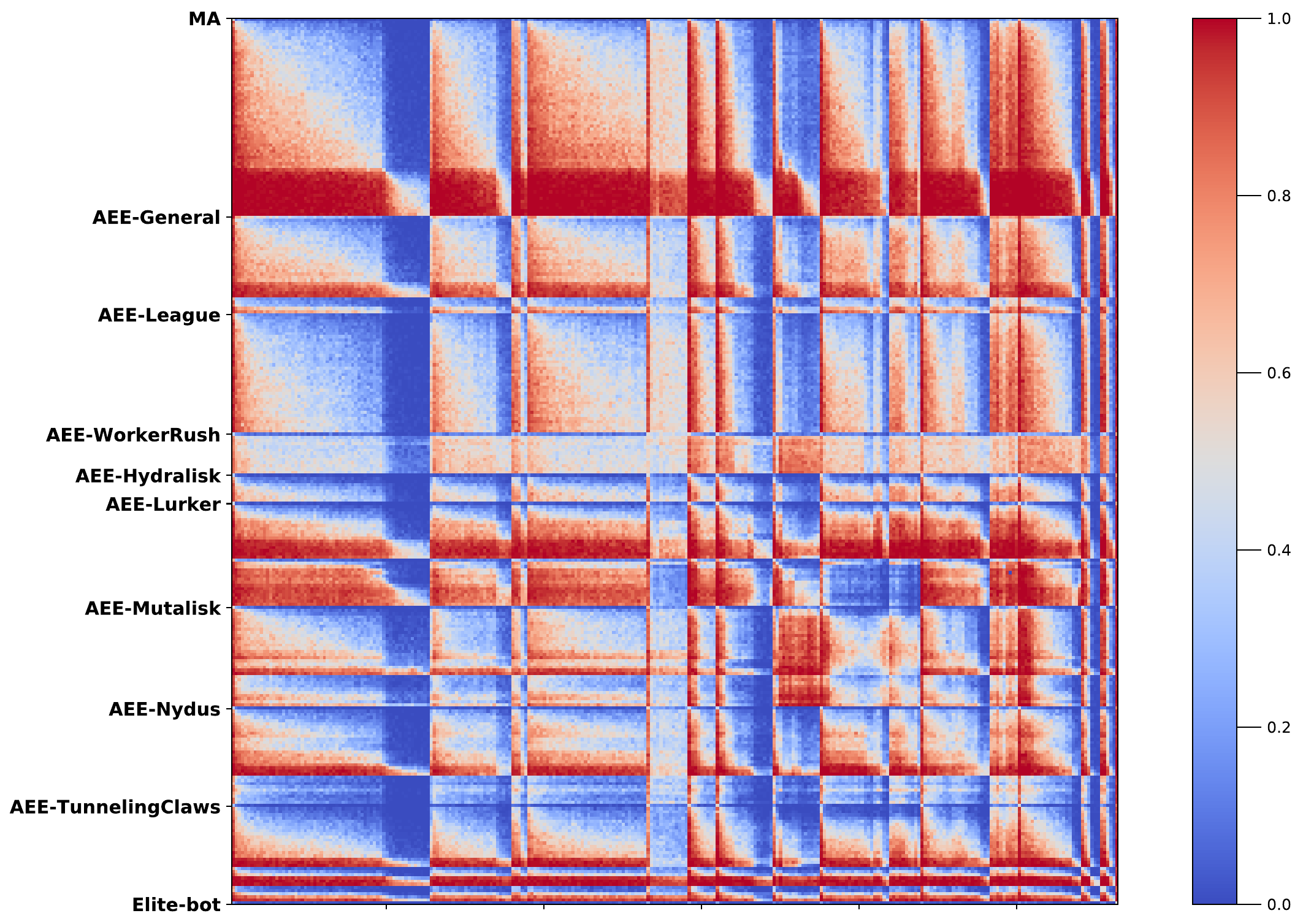}
\label{fig:payoff}
}
\subfigure[]{
\includegraphics[width=0.45\linewidth]{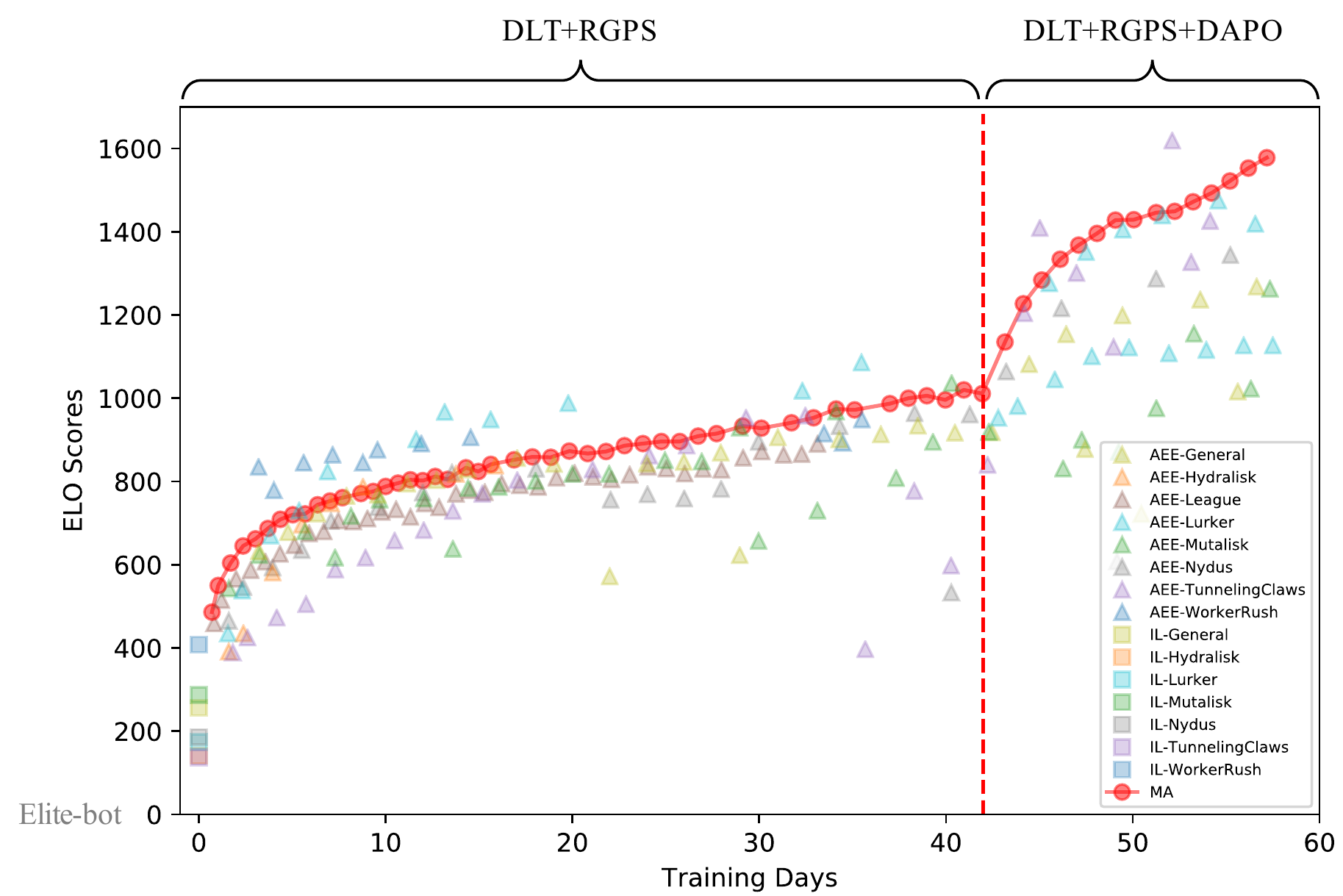}
\label{fig:elo}
}
\caption{
(a) The payoff matrix from the Round Robin tournament of the models in the formal experiment's league.
(b) ELO scores of agents in the formal experiment's league. 
Each point represents a historical agent, evaluated against the entire league and the Elite-bot, 
which is referenced as the baseline with zero ELO score. Cycles indicate MAs, triangles indicate
exploiters with their types in different colors, and squares are supervised models.
}
\end{figure}

In the formal experiment, the league consists of one MA and 8 exploiters. The MA is assigned with 32 GPUs for training and each exploiter is
assigned with 8 GPUs for training, and hence there are a total number of 96 GPUs used for training and another 48 GPUs used for 
inference via InfServer. The 8 exploiters are combinations of ME/LE/SE and AEE, i.e., AEE with distinct initial
supervised model or opponent matching mechanism, which we denote as AEE-General (ME+AEE), AEE-League (LE+AEE), 
AEE-Hydralisk (SE+AEE), AEE-Lurker (SE+AEE), AEE-Mutalisk (SE+AEE), 
AEE-Nydus (SE+AEE), AEE-TunnelingClaws (SE+AEE) and AEE-WorkerRush (SE+AEE).
In the first 42 days of training, DLT and RGPS are activated, and after that DAPO is activated 
additionally. 
This is an empirical choice, and it is expensive to repeat the entire formal experiment with different configurations.
However, we will use some shorter league training experiments to evaluate the effectiveness of each component.
Table~\ref{tab:rules} provides a list of all rules used in RGPS in our experiments.
\begin{table}[t]
  \centering
  \small
  \caption{Rules used in RGPS.}
  \label{tab:rules}
  \begin{tabular}{c|c}
    \hline
    \hline
    Rules on & Explanation \\
    \hline
    Morph Lair & After observing opponent has done so \\
	Morph Overseer & After observing opponent's Lurker \\
	Build Lurkerden & After observing opponent's Lurker \\
	Build Spire/Hydraliskden & After observing opponent's air unit \\
	Train Viper & After seeing many opponent's Lurkers \\
	Morph Hive & If `Train Viper' rule is activated \\
    \hline
    \hline
  \end{tabular}
\end{table}

During 57 days of training, a total number of 25,708,491 matches are generated
by all the actors, and 
a total number of 583 agents are generated, including 124 main agents, 452 exploiters and 
7 initial supervised models. We select half of these models stored every two learning periods, and 
perform Round Robin tournament among these models. That is, any pair of these models will be tested for 
100 matches. For all the testing matches, the configuration of $z$-stat follows the configuration in training. 
After a total number of around 4 million testing matches, we obtain the payoff matrix shown in
Figure~\ref{fig:payoff}. The rows/columns are ranked by agent types and training time. Important
conclusions can be drawn from Figure~\ref{fig:payoff} are: (1) for each type of the agents, their
strength is consistently increasing with higher win-rates against earlier models; (2) for each 
type of the exploiters, continually training allows they are consistently competitive 
compared to MA (recall the working mechanism of AEE); by comparision, the payoff 
matrix in AlphaStar~\cite{astar} shows that its MEs can rarely catch up with its MA as the training
proceeds, and therefore these computational efforts should be more reasonably reallocated; (3) 
the strength of all the agents is significantly improved after DAPO activated. 
Figure~\ref{fig:elo} shows the ELO scores of the agents in the league. As we can observe, the MA
consistently improves while the exploiters can also catch up with the MA as the training proceeds.
Again, it is obvious that DAPO significantly promotes the policy improvement after its activation.

\begin{figure}[t]
\center
\subfigure[]{
\includegraphics[width=0.45\linewidth]{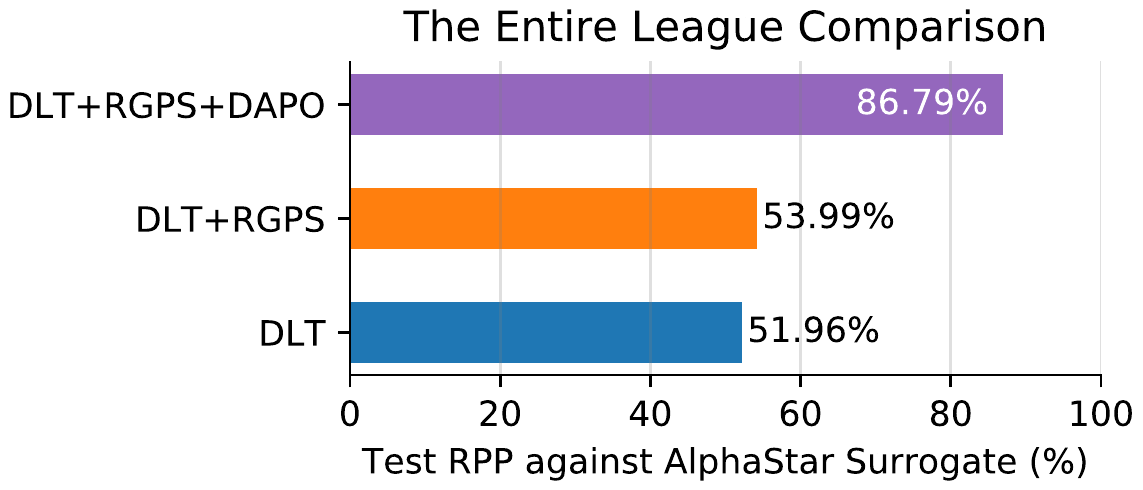}
\label{fig:rpp_league}
}
\subfigure[]{
\includegraphics[width=0.45\linewidth]{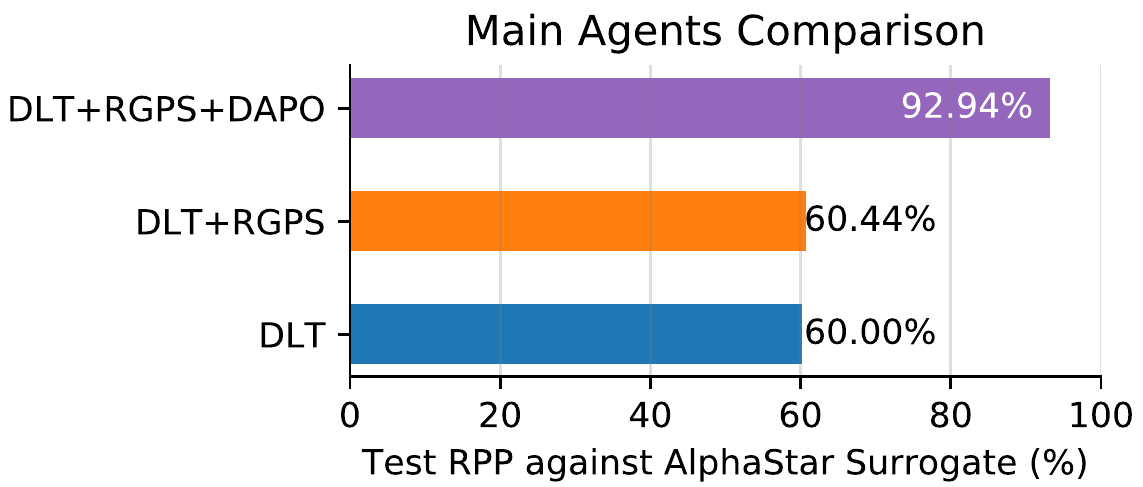}
\label{fig:rpp_ma}
}
\caption{
Comparing different league compositions using RPP. (a) comparison between the entire leagues; (b) comparison between MAs.
}
\label{fig:rpp}
\end{figure}

\subsubsection{Key Components Evaluation in League Training}

\begin{figure}[t]
\center
\subfigure[]{
\includegraphics[width=0.45\linewidth]{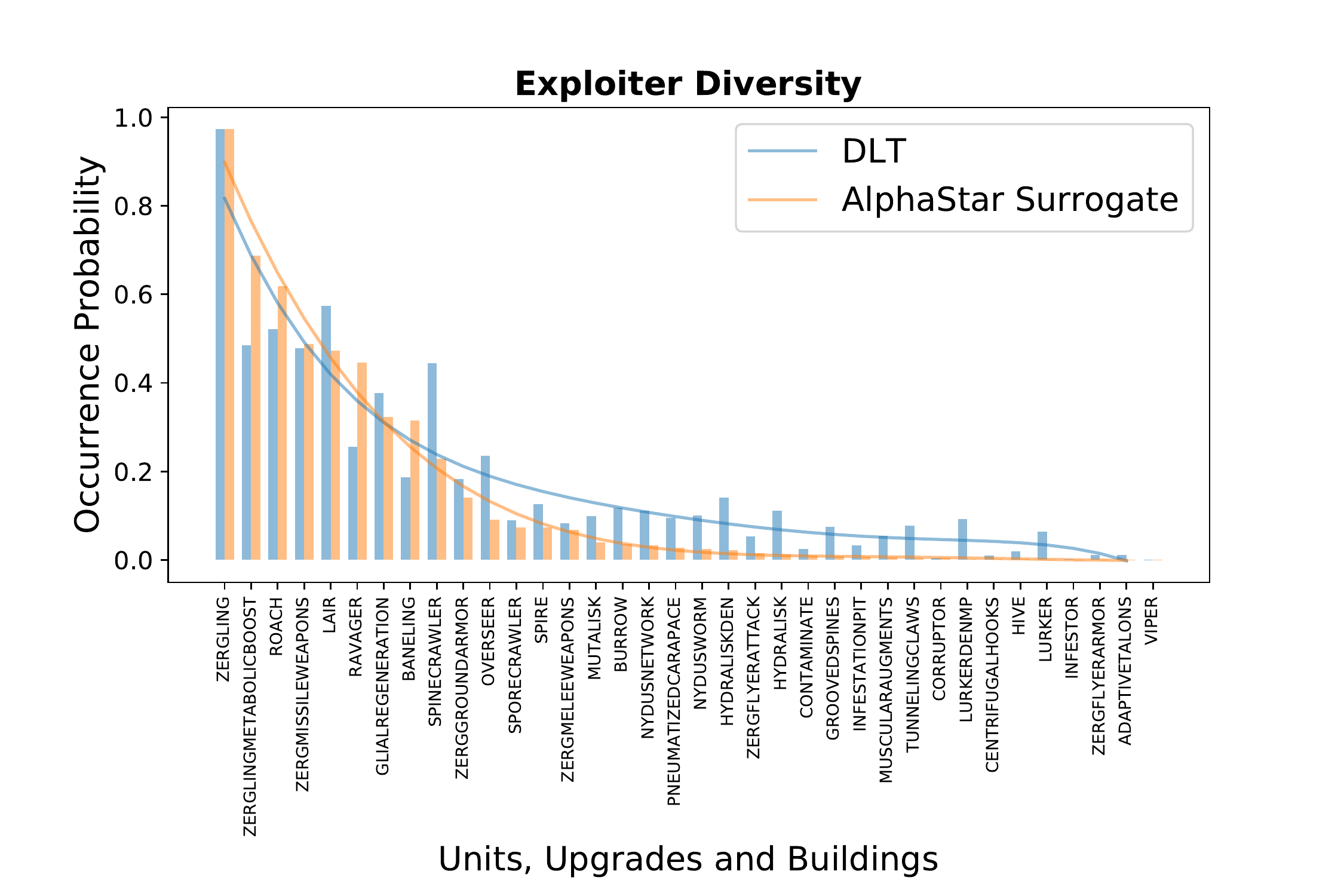}
\label{fig:diversity_oppo}
}
\subfigure[]{
\includegraphics[width=0.45\linewidth]{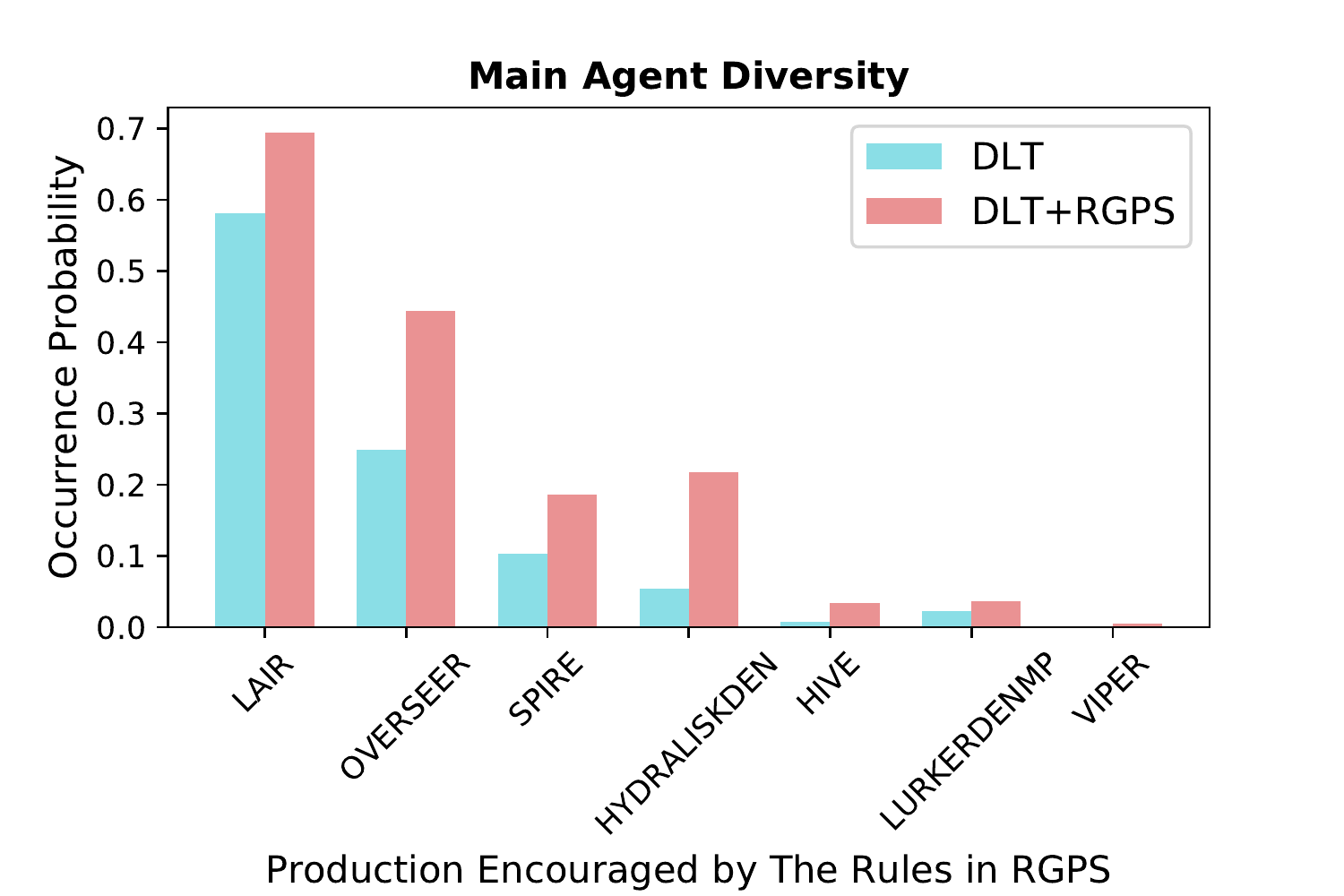}
\label{fig:diversity_ma}
}
\caption{
(a) Exploiters' diversity in the league: 
units, upgrades and buildings' occurrence probability in all exploiters' 
matches of last training day of the 7 days.
(b) The occurrence probability of the units and buildings considered in RGPS with respect to Table~\ref{tab:rules}, counted in MA's matches in the last training day of the 7 days.
}
\end{figure}

In addition to the formal experiment, we also conduct several ablation studies, each of which uses full scale of the computational
resources as the formal experiment but runs for a week. The compared settings are AlphaStar Surrogate, DLT (only), DLT+RGPS and DLT+RGPS+DAPO.
AlphaStar Surrogate is constructed by composing the league with one MA (assigned with 32 GPUs), two MEs (each with 16 GPUs), and two LEs 
(each with 16 GPUs). To compare among different leagues, we use the Relative Population Performance (RPP)~\cite{astar}, that is the
payoff value of the Nash solution between two leagues. To test the RPP value between league A and league B, we let every agent in A play
against every agent in B with 100 matches. For all the testing matches, the configuration of $z$-stat follows the configuration in training.
We use the AlphaStar Surrogate as the reference league. Figure~\ref{fig:rpp} shows
the test RPP of different configurations when comparing the whole league and the MAs, respectively. By using DLT, although the entire league's
RPP is only slightly higher than AlphaStar Surrogate, the RPP of the MAs increases apparently, indicating the MAs in DLT are more robust
and stronger than the MAs in AlphaStar Surrogate. It is not surprising that the RPP increment of the entire league of DLT is not obvious, because
at early stage of the league training, MA is not strong and AEEs can easily catch up with MA; then, according to the scheme of AEE, these
exploiters reset to their supervised agents, and therefore act like ME and LE; also, in DLT, each exploiter is assigned with only 8 GPUs and hence
their policy improvement is limited compared to MEs and LEs in AlphaStar Surrogate. However, the diversity generated from DLT results in more robust MAs. 
This can further be verified in Figure~\ref{fig:diversity_oppo}, which shows the occurrence 
probability of most main Zerg units, upgrades and buildings in all the training matches of the 
last day of training. From the figure, it is clear that the exploiters in DLT show much diverse
units, upgrades and buildings, while the non-zero quantities of AlphaStar Surrogate are mainly
concentrated at the left side. By adding RGPS, we also observe slight RPP improvement of both
the league and MAs comparison, with DLT as the reference. It is not surprising either that the
RPP values are with slight increment because very few critical states are considered in RGPS 
(see Table~\ref{tab:rules}). However, when considering these critical states, 
Figure~\ref{fig:diversity_ma} demonstrates that RGPS indeed fits these rules in the policy to
enhance the probability of these relevant actions.
Finally, after adding DAPO, the RPP values against AlphaStar Surrogate significantly increases for 
both league and MAs comparisons, showing the effectiveness of DAPO in promoting policy 
improvement.

\subsection{Other Results}
\subsubsection{Build-in Bots Evaluation}
\begin{figure}[t]
\center
\subfigure[W/ importance sampling]{
\includegraphics[width=0.3\linewidth]{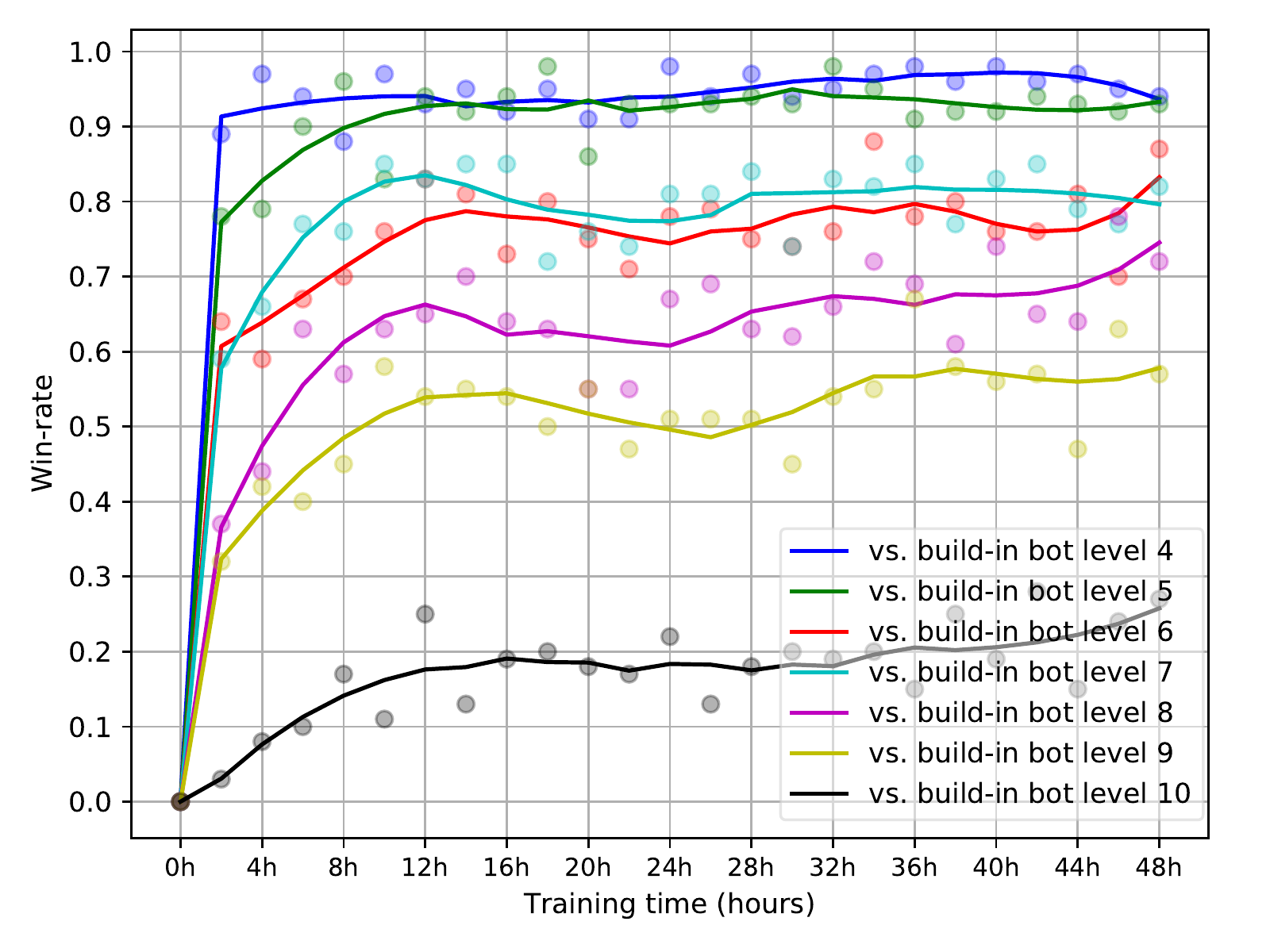}
\label{fig:il}
}\hskip -0.1in
\subfigure[W/o importance sampling]{
\includegraphics[width=0.3\linewidth]{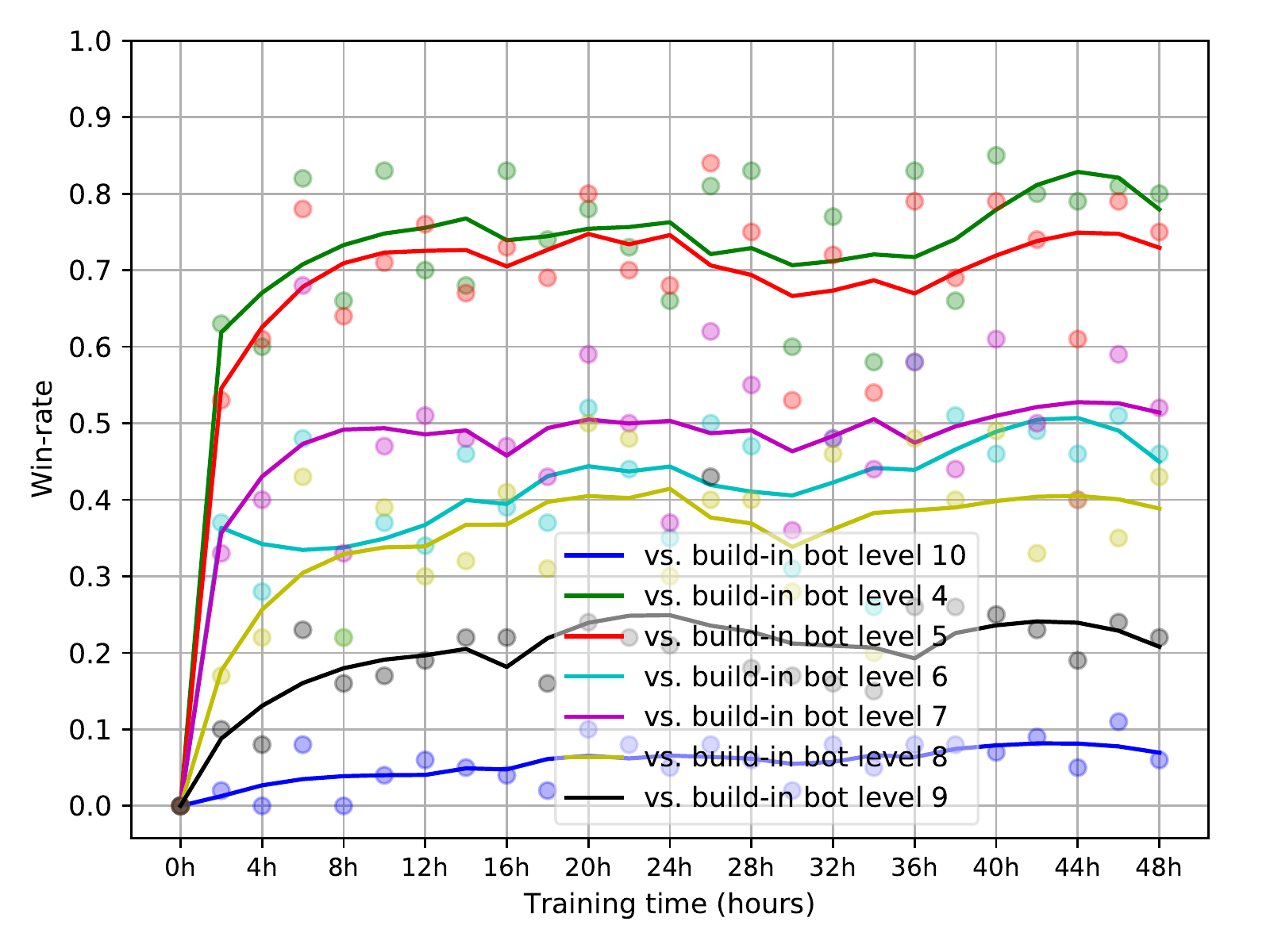}
\label{fig:il-nois}
}
\subfigure[Win-rates in early league]{
\includegraphics[width=0.3\linewidth]{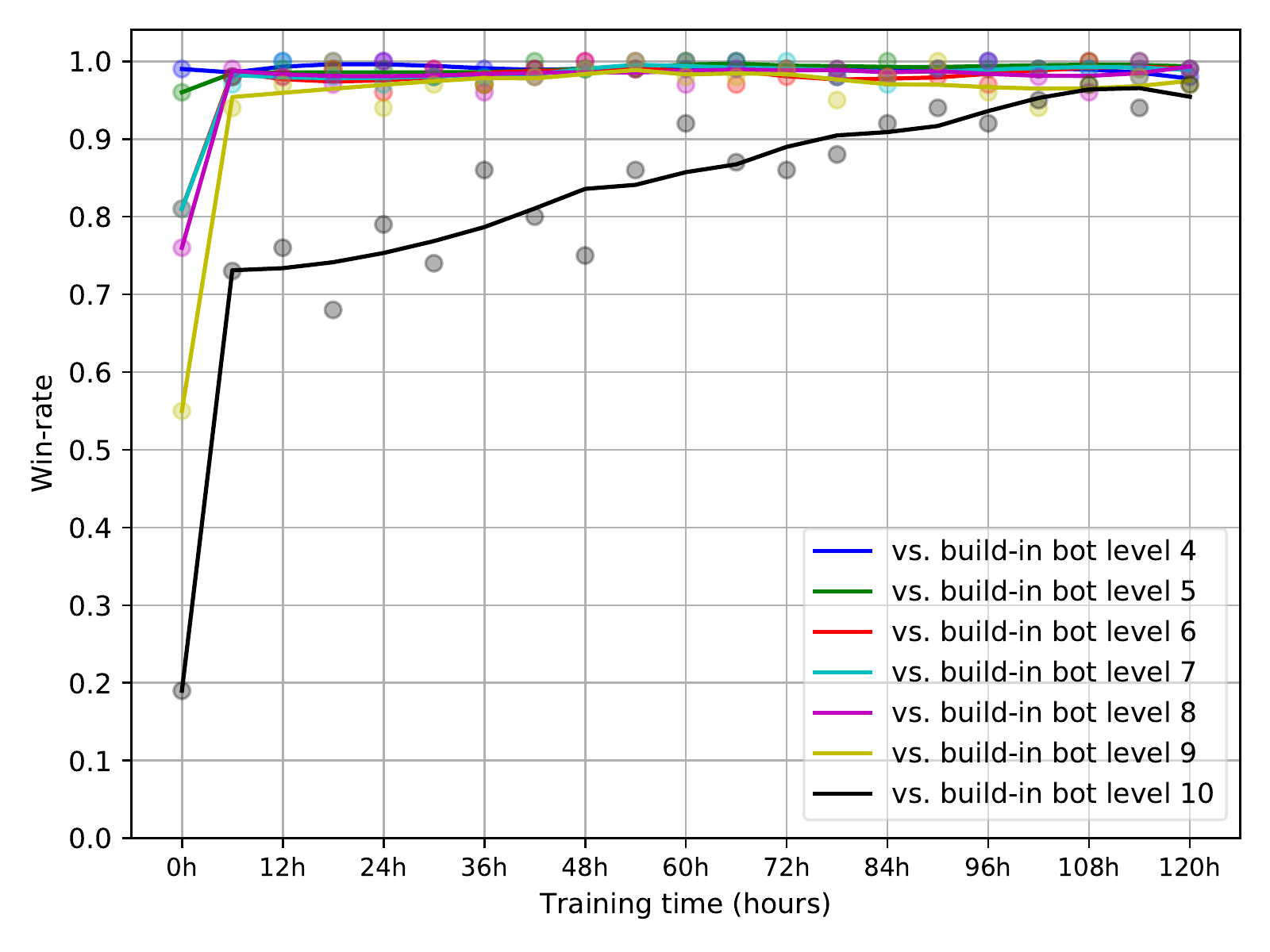}
\label{fig:rl}
}
\caption{(a)-(b) Comparison of IL with and without importance sampling. (c) Win-rates of the agents in early league training. Each point indicates the win-rate of a stored model against the corresponding build-in bot, averaged over 100 testing matches.}
\end{figure}

We report build-in bots evaluation results for both the supervised agents from IL and some early
agents from the formal experiment.
Figure~\ref{fig:il} shows the testing win-rates of the supervised agents against 
the build-in bots from level 4 to level 10 as the imitation learning proceeds. 
Each point indicates an average win-rate from 100 matches and the agent randomly samples 
a $z$-stat from a subset of 174 human replays with the winner's MMR $\geq$ 6200 on the 
Kairos Junction map. For an ablation study, we also provide the training curve by removing 
importance sampling and Figure~\ref{fig:il-nois} shows the test win-rates.
Obviously, training the agent using the original data distribution significantly reduces 
the performance.
Similar to IL, we provide evaluation results of the main agents from early league training against the build-in bots. 
Figure~\ref{fig:roles}
shows the testing curve and we observe that the reinforced agent can consistently enhance its performance against
the build-in bots although it never meet any of them in its league training. 





\begin{figure*}[!ht]
\center
\includegraphics[width=1.0\linewidth]{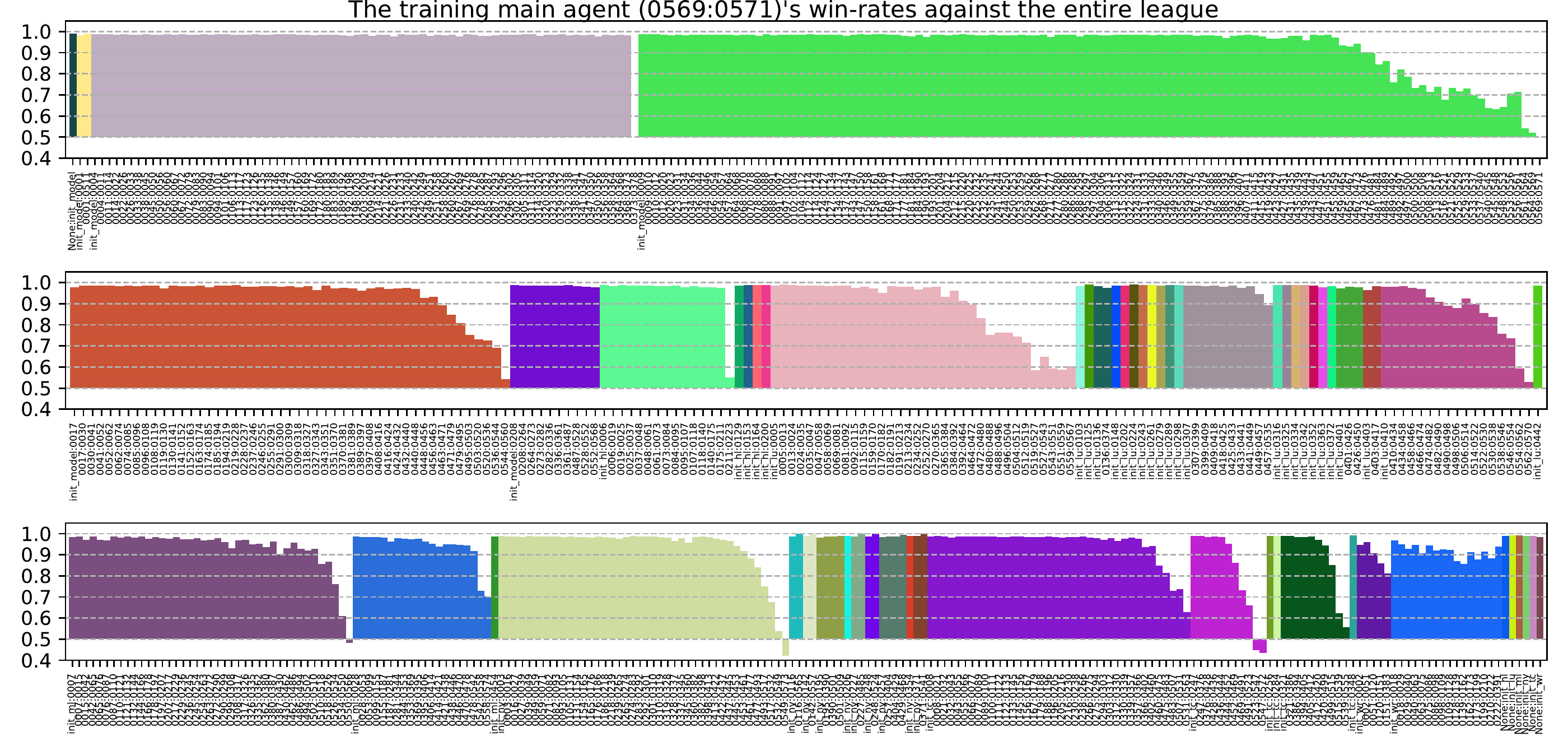}
\caption{The training MA's win-rates against the entire league at the 57th day of training.}
\label{fig:league-bar}
\end{figure*}


\subsubsection{Opponent Matching Status in League Training}
Figure~\ref{fig:league-bar} shows the opponent matching status of MA in the entire league on the final day 
of the formal experiment. A number of 583 agents are generated and each agent is denoted by a bar in
the Figure. The agent ID consists of a parent ID following by the descendent ID, 
separated by a colon. If the parent ID is `None', it indicates that the model is one of the initial models.
The bar values indicate the win-rates of the training MA against all other agents in the league at that time. 
The group of green bars (right hand side of the top sub-figure) indicates all the MAs generated in history
and the right-most one is the MA that is under training. All the other colorful bars indicate various
exploiters. Groups of bars located together with the same color indicate an evolving branch of agents. 
As we can observe, the training MA can almost beat all the historical agents with win-rate above 
50\% except a few most recent exploiters.




\subsubsection{Diversity in Supervised Models and MA's Counter Strategies}

\begin{figure}[t]
\center
\includegraphics[width=1.0\linewidth]{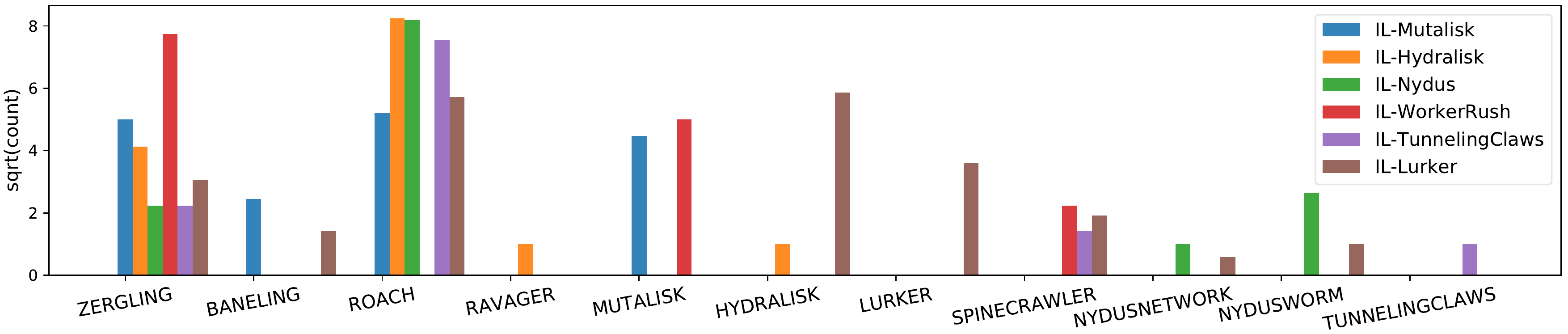}
\caption{Average (square root) unit counts for the fine-tuned IL models in testing matches against the Elite-bot.}
\label{fig:6ft-uc}
\end{figure}

In IL, we fine-tuned $K=6$ supervised models on 6 subsets of human replays, each of which includes
one the following units/upgrades/strategies: Hydralisk, Lurker, Mutalisk, Nydus Network, 
TunnelingClaws, and Worker Rush. We indeed tried more subsets from common human strategies 
while we find that the above selected strategies are relatively more competitive.
We evaluate each of them by 100 matches versus the Elite-bot and the collected unit
counts are shown in Figure~\ref{fig:6ft-uc}. The results demonstrate that 
fine-tuning the model using distinct subsets 
of replays is empirically effective to provide initial diversities.
\begin{figure}[t]
\center
\includegraphics[width=1.0\linewidth]{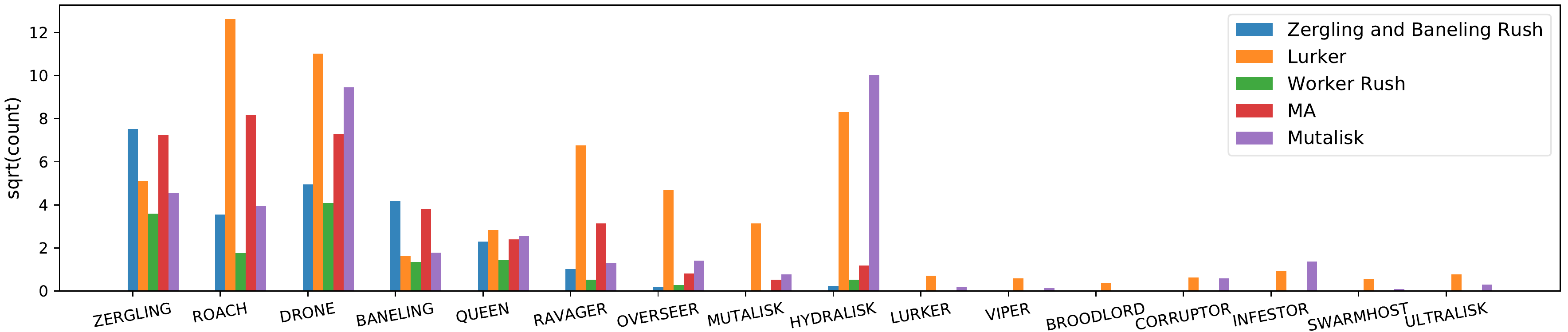}
\caption{Average (square root) unit counts in MA's counter strategies when it plays against diverse opponents.}
\label{fig:counter}
\end{figure}
Next, we show the robustness of the MA. A simple way to access MA's robustness is varying its opponents with 
diverse strategies and see how MA responses. For demonstration, we choose some representative exploiters 
each of which focusing on a distinct strategy, including using Zergling and Baneling rush, Worker rush, 
Mutalisk and Lurker, respectively. We then let MA (with zero $z$-stat) 
play against these exploiters and itself to see how MA reacts after seeing
different strategies. Figure~\ref{fig:counter} shows the average unit counts in MA's counter strategies 
when it plays against diverse opponents (100 matches for each). 
As we can see, MA performs multiple strategies when it encounters different opponents. 
When its opponent uses Zergling and Baneling Rush, it 
responses with Zerglings and Banelings as well; 
when its opponent produces Lurkers, the MA tends to produce more
Overseers; when its opponent uses Mutalisk, it bursts a large number of Hydralisks; for Self-Play, it prefers
to use Roach, Zergling and Ravager. These results demonstrate that the unconditional MA is a robust agent and is able to use
multiple counter-strategies.

\subsubsection{Infrastructure Comparison}
We use a distributed learner-actor infrastructure with 144 Tesla V100 GPUs and 13,440 CPU cores. 
In our formal experiment, 96 GPUs are used for training and the rest GPUs are used for inference. 
Generally, we assign 32 GPUs for training the MA and 64 GPUs are alternatively allocated to different exploiters. 
Each training GPU is allocated with 70 actors, running 70 concurrent matches, and each actor is assigned with 2 CPU cores.
A total number of used GPUs and CPUs has been given in Table~\ref{tab:com}. All the computation resources are 
deployed on the Tencent Cloud. In our infrastructure, the data consuming speed is 210 fps/GPU, and the replay buffer
of each GPU receives data with a speed of 43 fps. That is, each data point will be used five times in average.
In AlphaStar's infrastructure, its consuming speed is 390 fps/TPU core and the data generating speed is 195 fps with
each data sample used twice in its replay buffer. Table~\ref{tab:com} provides a detailed comparison.
From the perspective of the entire league training, AlphaStar's total consuming/generating speed in the league is about 
30/73 times faster than that in TStarBot-X. Although this is indeed not straightforwardly comparable, since 
AlphaStar learns to play three races on four SC2 maps and we only focus on Zerg vs. Zerg on the Kairos Junction map,
we can still observe the scale differences of the two systems.

\section{Discussion and Conclusion}
\label{sec:conclusion}

\begin{figure}[t]
\center
\includegraphics[width=1.0\linewidth]{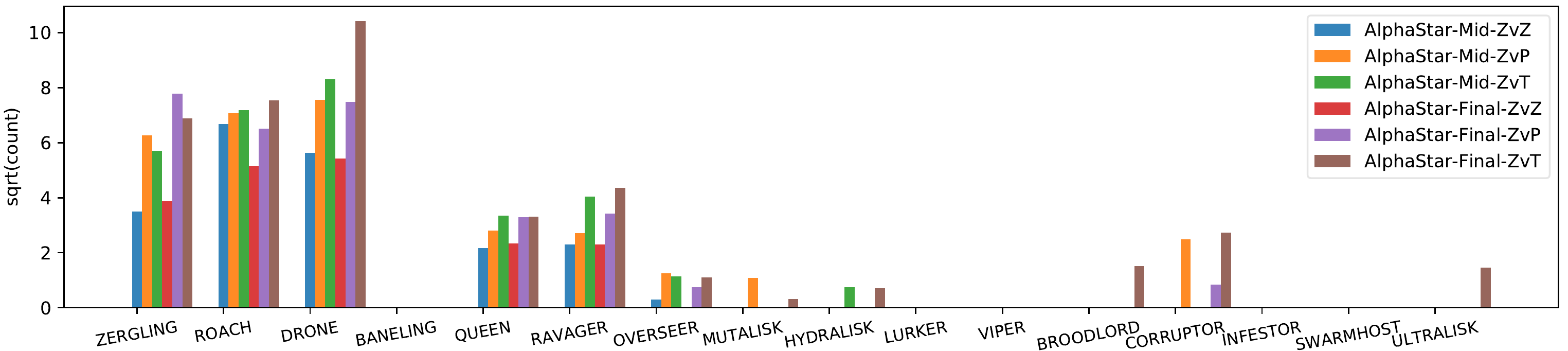}
\caption{Average (square root) unit counts in AlphaStar Zerg's human evaluation replays.}
\label{fig:astar-uc}
\end{figure}

We end up the manuscript with a discussion of the AlphaStar and TStarBot-X's final agents. 
With curiosity on the Zerg MA's performance in AlphaStar, we parse all AlphaStar's human evaluation replays and show its 
unit counts of the Zerg agent in Figure~\ref{fig:astar-uc}. It seems that AlphaStar's Zerg agent prefers to use a 
specific strategy that only produces Roach, Zergling, Ravager and Queen in all its Zerg vs. Zerg (ZvZ) matches. 
A few other units can be observed only in its Zerg vs. Terran (ZvT) and Zerg vs. Protoss (ZvP) matches. 
Although these statistics might be biased from AlphaStar's training distribution, it probably suggests that 
1) exploration and keeping diversity (in a single policy) in ZvZ game are difficult;
2) a joint training with all three races can enhance the diversity in the league.
TStarBot-X has a more diversified league and more strategic diversity in a single policy, even though it is trained
under a much smaller scale of computations and ZvZ setting. However, TStarBot-X is not good at multitasking and 
counter-multitasking, from the evaluation of Grandmaster human player.

Overall, in this paper, we provided a detailed study on training a competitive AI agent in 
SC2 under limited scale of computation resources. 
We encountered problems when we reimplemented the settings in AlphaStar, and we then proposed new
techniques to solve these problems. The study suggests that 1) when the computations are insufficient
to afford enough explorations in a complex problem, diversified league training with mixture of the proposed agent roles 
could be a more efficient way to provide diversity in the league; 2) rule-guided policy search is an effective way to fuse
human knowledge into a neural network to avoid tedious explorations; 3) divergence-augmented policy optimization
on the agent level can significantly enhance policy improvement. 
Most importantly, the entire system is open-sourced and we expect that this might be useful for future 
AI research in solving similar problems, policy transfer from the released parameters, and so on. For future directions, 
we are interested in applying the proposed techniques and infrastructure to other complex multi-agent systems and apply to
read-world applications.

\section*{Acknowledgement}
We would like to thank Zhong Fan, Lei Jiang, Yan Wang, Zhihuai Wen and Yang Liu for providing us 
specialized advices in SC2. We would like
to thank our colleagues from Department of AI Platform and Tencent Cloud for their valuable discussions and supports.
We would like to show respect to and thank Blizzard Entertainment for creating StarCraft. 
Lei Han and Peng Sun are faithful players of StarCraft: Brood War since its release. Lei Han plays Zerg. 
Peng Sun and Lei Han would like to show gratitudes to their boyhood friends who 
had ever played StarCraft together with them. 
We could not play StarCraft forever but can be brothers for a lifetime.

\section*{Author Contributions}
Lei Han, Jiechao Xiong and Peng Sun contributed equally. Peng Sun organized the infrastructure. 
Lei Han, Jiechao Xiong and Peng Sun designed, built and experimented 
multiple versions of environmental spaces, 
neural network architectures, RL algorithms, league training methods, etc.,
with early contributions from Xinghai Sun. 
Peng Sun, Jiechao Xiong and Lei Han completed the engineering, evaluation, visualization 
and analyzing tools. 
Meng Fang, Qingwei Guo, Qiaobo Chen, Tengfei Shi, Hongsheng Yu and Xipeng Wu helped 
reviewing the project codes and experimenting with some ablations. 
Lei Han wrote the paper with contributions from Peng Sun, Jiechao Xiong and Meng Fang.
Zhengyou Zhang provided general scope advices and consistently supported the team.

\small

\bibliography{cite}
\bibliographystyle{unsrt}


\end{document}